\pdfoutput=1
\documentclass[fleqn]{article}
\usepackage{arxiv/arxiv}

\usepackage{authblk}
\usepackage[square,numbers]{natbib}
\usepackage[utf8]{inputenc} %
\usepackage[T1]{fontenc}    %
\usepackage{hyperref}       %
\usepackage{url}            %
\usepackage{booktabs}       %
\usepackage{amsfonts}       %
\usepackage{nicefrac}       %
\usepackage{microtype}      %

\usepackage[dvipsnames]{xcolor}         %
\usepackage{nicefrac}       %
\usepackage{array}
\usepackage{graphicx}
\usepackage{wrapfig}  
\usepackage{enumitem}       %
\usepackage{amsmath}
\usepackage{setspace}
\usepackage{makecell}
\usepackage{tabularx}

\makeatletter
\renewcommand\AB@authnote[1]{\textsuperscript{\normalfont\bfseries#1}}
\makeatother

\newcommand{\vparagraph}[1]{\vspace{-2.7mm}\paragraph{#1}}

\newcommand{\vsection}[1]{\vspace{-1mm}\section{#1}\vspace{-1mm}}

\newcommand{\vsubsection}[1]{\vspace{-1mm}\subsection{#1}\vspace{-1mm}}

\let\cite\citep

\newcommand{\authemail}{\textsuperscript{$\dagger$}}

\newcommand{\textscript}[1]{\text{\scriptsize ~#1}}

\title{Reward (Mis)design for Autonomous Driving}

\author[1,2]{W. Bradley Knox\thanks{Corresponding author: \texttt{bradknox@cs.utexas.edu}}~~\textsuperscript{,}}
\author[1,2]{Alessandro Allievi}
\author[3]{Holger Banzhaf\,}
\author[4]{Felix Schmitt}
\author[2,5]{Peter Stone}
\affil[1]{Robert Bosch LLC}
\affil[2]{The University of Texas at Austin}
\affil[3]{Robert Bosch GmbH}
\affil[4]{Bosch Center for Artificial Intelligence}
\affil[5]{Sony AI}

\begin{document}

\maketitle

\begin{abstract}

This article considers the problem of diagnosing certain common errors in reward design. Its insights are also applicable to the design of cost functions and performance metrics more generally. 
To diagnose common errors, we develop 8 simple sanity checks for identifying flaws in reward functions. These sanity checks are applied to reward functions from past work on reinforcement learning (RL) for autonomous driving (AD), revealing near-universal flaws in reward design for AD that might also exist pervasively across reward design for other tasks. Lastly, we explore promising directions that may aid the design of reward functions for AD in subsequent research, following a process of inquiry that can be adapted to other domains.
\end{abstract}

\vsection{Introduction}

Treatments of reinforcement learning often assume the reward function is given and fixed. However, in practice, the correct reward function for a sequential decision-making problem is rarely clear. Unfortunately, the process for designing a reward function (i.e., \emph{reward design})---despite its criticality in specifying the problem to be solved---is given scant attention in introductory texts.\footnote{Unless otherwise noted, any discussion herein of reward design focuses on the specification of the environmental reward, \emph{before any shaping rewards are added}. We also focus by default on \textit{manual} reward specification, which differs from inverse reinforcement learning and other methods for learning reward functions. However,  we discuss the application of this work to such methods in Section~\ref{sec:rewlearning}.} For example, \citeauthor{sutton2018reinforcement}'s standard text on reinforcement learning~\citeyearpar[p.~53--54, 469]{sutton2018reinforcement} devotes merely 4 paragraphs to reward design in the absence of a known performance metric. Anecdotally, reward design is widely acknowledged as a difficult task, especially for people without considerable experience doing so. %
Further, \citet{dulac2021challenges} recently highlighted learning from ``multi-objective or poorly specified reward functions'' as a critical obstacle hampering the application of reinforcement learning to real-world problems.
Additionally, the problem of reward design is highly related to the more general problem of designing performance metrics for optimization---whether manual or automated optimization---and is equivalent to designing cost functions for planning and control (Section~\ref{sec:background}), making a discussion of reward design relevant beyond reinforcement learning. This article contributes to an important step in reward design: evaluating a proposed reward function independently of which reinforcement learning algorithms are chosen to optimize with respect to it. To this end, in Section~\ref{sec:tools} we develop 8 sanity checks for identifying flaws in reward functions. 

Throughout this article, we use autonomous driving as a motivating example for reward design. AD also serves as a source of reward functions to demonstrate application of our 8 sanity checks. In Section~\ref{sec:challenge}, we describe challenges of reward design for autonomous driving. 
The sanity checks in Section~\ref{sec:tools} reveal pervasive issues in published reward functions for AD. Specifically, the majority of these reward functions---often all of them---fail the tests that are part of the first 3 sanity checks (Sections~\ref{sec:shaping}--\ref{sec:indifference}). In Section~\ref{sec:design}, we further explore obstacles to reward design for AD through initial attempts to design three attributes of an AD reward function, uncovering obstacles to doing so that a designer should consider. Section~\ref{sec:design} also includes a review of some government-mandated performance metrics, discussions of reward \emph{learning} for AD and of multi-objective optimization for AD, and a proposal for designing reward for AD with a financial currency as its unit.
Given the high envisioned impact of autonomous driving, the RL-for-AD community needs to consider reward design carefully if reinforcement learning will have a significant role in the development of AD.

We do not seek to condemn the past efforts we review---which are excellent in many regards---but rather to provide understanding of common issues in reward design and guidance on identifying them. Additionally, this paper is not a general survey of reinforcement learning for autonomous driving; such surveys have been conducted by \citet{zhu2021survey} and \citet{kiran2021deep}.
The contributions of this article include a deep discussion of what reward design for AD should entail, the development of 8 simple sanity checks for reward or cost functions, application of these sanity checks to identify prevalent flaws in published reward functions for AD (flaws that anecdotally appear common throughout RL, beyond AD), identifying pervasive usage of \emph{trial-and-error reward design} (see Section~\ref{sec:redflags}), and revealing obstacles that arise in our initial attempt to design reward for AD. In particular, we do not claim to solve the problem of reward design for AD. Instead, we provide guidance for future efforts to design reward and other performance metrics for autonomous driving specifically as well as for other tasks with undefined reward functions.

\vspace{2mm}
\vsection{Background: objectives, utility functions, and reward functions}
\label{sec:background}

To support our discussion of reward design, we first review what a reward function is. When attempting to decide between alternative decision-making or control algorithms---or, equivalently in this discussion, \emph{policies}, which map each state to a probability distribution over actions---a common and intuitive approach is to define a \emph{performance metric} $J$ that scores each policy $\pi$ according to $J: \pi \rightarrow \mathbb{R}$. If $J(\pi_A) > J(\pi_B)$, then $\pi_A$ is better according to $J$.\footnote{Roughly speaking, many learning algorithms for sequential-decision making tasks can be thought of as looping over two steps. (1) Policy evaluation: approximately evaluate $J(\pi)$ for one or more policies via gathered experience. (2) Policy improvement: use the evaluation of step 1 to choose a new policy $\pi$ or several new policies to evaluate.}

Definition of the performance metric $J$ creates a ranking over policies and identifies the set of optimal policies, both of which are helpful for optimizing behavior in sequential decision-making tasks. However, different definitions of the performance metric typically create different rankings and sets of optimal policies, which generally change the result of a learning algorithm optimizing according to some $J$. \emph{Put simply, bad design of a performance metric function creates misalignment between what the designer---or other stakeholders---considers to be good behavior and what the learning algorithm does. A perfect optimization algorithm\footnote{We use ``learning'' and ``optimization'' interchangeably in this article.} is only as good as the performance metric it is optimizing.}

$J(\pi)$ is typically defined as the expectation over outcomes created by $\pi$, where these outcomes can be directly observed but $J(\pi)$ cannot. More specifically, in episodic framings of tasks, $J(\pi)$ is defined as an expectation of $G(\tau)$, which is the performance
of trajectory $\tau$ created by the policy. Specifically, $J(\pi) = E_{\pi,D} [G(\tau)]$, where $\tau$ is a trajectory generated by following $\pi$ from a starting state drawn according to some initial state distribution $D$. The mean of $G(\tau)$ from trajectories generated therefore can serve as a practical, unbiased estimator of $J(\pi)$.\footnote{Episodic task framings are the norm in reinforcement learning for AD and are assumed throughout this article unless otherwise noted.}

\begin{wrapfigure}{}{0.65\textwidth}
    \hspace{0mm}
    \centering
    \includegraphics[width=0.9\linewidth]{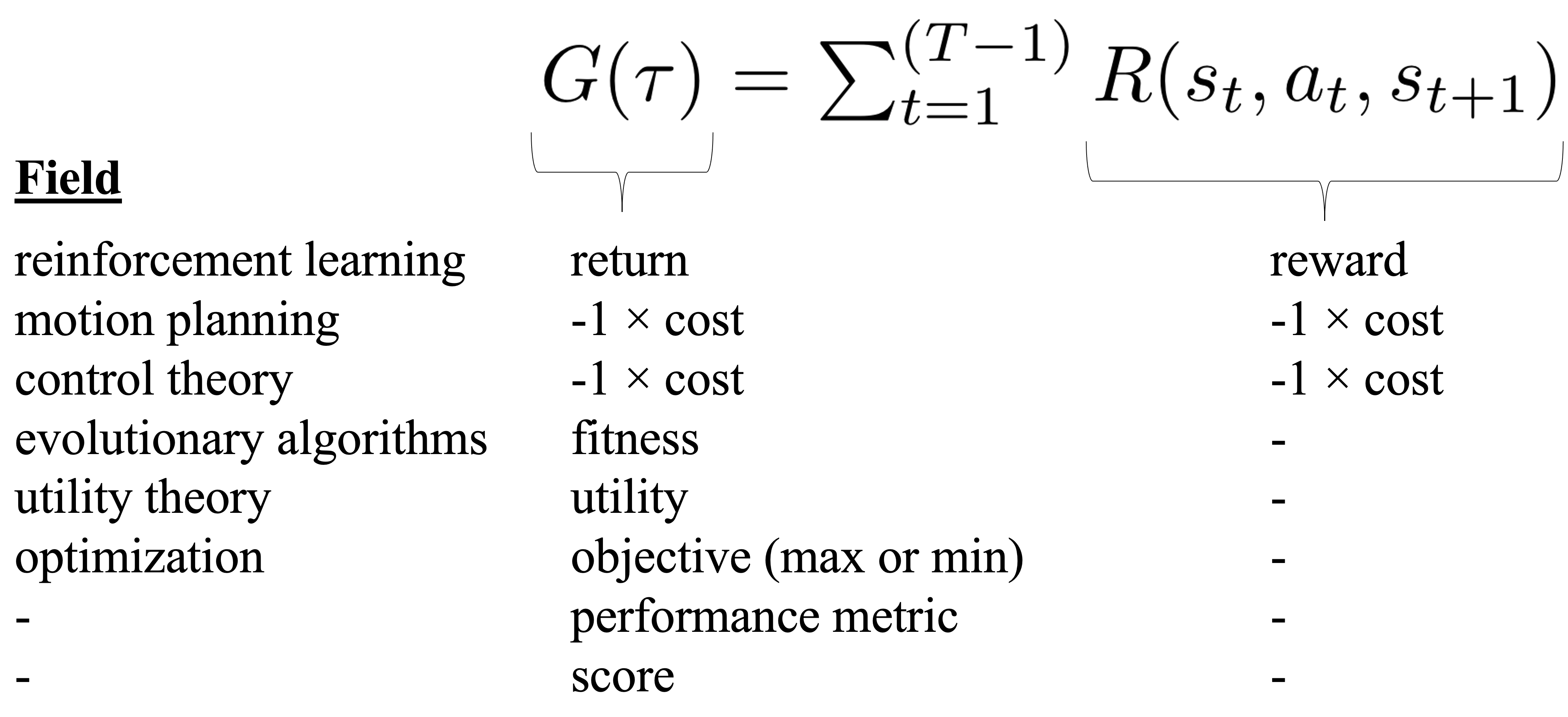}
    \vspace{0mm}
    \caption{\small{Relationships of terminology among related fields. Note that \emph{fitness} and \emph{objective} are similar in meaning to \emph{return} but not identical, as we describe in Section~\ref{sec:background}.}}
    \vspace{0mm}
    \label{fig:terminology}
\end{wrapfigure}

The policy-performance metric $J$ and the \emph{trajectory-performance metric} $G$ go by many names in the various fields that address the sequential decision-making problem. Figure~\ref{fig:terminology} shows some of these various terms. In many subfields, $J$ is called an \emph{objective} if its definition explicitly includes the goal of maximizing it (or of minimizing it, in some contexts). In utility theory, $G$ is a \emph{utility function} and $J(\pi)$ is the \emph{expected utility}. In planning and control theory, both $J(\pi)$ and $G(\tau)$ can be referred to as \emph{cost} (and optimization seeks to minimize cost rather than maximize it, as we otherwise assume here). In evolutionary algorithms, $J(\pi)$ is referred to as the \emph{fitness} of a policy. Fitness may simply be the mean of $G(\tau)$ over $n$ samples of $\tau$, not the expectation $G(\tau)$. In \emph{reinforcement learning (RL)}, each transition within the trajectory elicits a \emph{reward} signal according to a \emph{reward function} $R : S \times A \times S \rightarrow \mathbb{R}$. The input for $R$ is the state, action, and next state in the transition. In the undiscounted setting, the sum of a trajectory $\tau$'s rewards is its \emph{return}, which is reinforcement learning's word for $G(\tau)$. Therefore, $G(\tau) = \sum_{t=1}^{(T-1)} R(s_t,a_t,s_{t+1})$, where trajectory $\tau = (s_1, a_1, s_2, a_2, ..., s_{T-1}, a_{T-1}, s_{T})$. Policy-improvement algorithms in RL seek to maximize the expected return, $J(\pi)$.%

We will largely use the terminology of RL, but we will refer to trajectory-level performance metrics as utility functions, since ``return function'' is not a common concept. Also, where feasible, this article's discussions focus on trajectory-level utility functions, $G$s, rather than reward functions. We do so for clarity, having judged subjectively that the consequences of a utility function are more analytically accessible than those of a reward function.

\vsection{The challenge of reward design for autonomous driving}
\label{sec:challenge}

We now consider four challenges of reward design for AD. We note however that these challenges apply widely to other tasks, particularly those that occur in the physical world and therefore can have effects beyond the typically envisioned scope of the task.

\vparagraph{Utility function depends on numerous attributes} First, driving is a \emph{multiattribute problem}, meaning that it encompasses numerous attributes that each contribute to the utility of driving. 
These attributes may include measurements of %
\textbf{progress to the destination, time spent driving, collisions, obeying the law, fuel consumption, vehicle wear, passenger experience, and various impacts on the world outside the vehicle}. These external impacts include those on people in other cars, pedestrians, bicyclists, and the government entity that builds and maintains driving infrastructure, as well as pollution and the climate more broadly. Defining a trajectory-performance metric $G$ for AD requires (1) identifying all such attributes and specifying them quantitatively and (2) combining them into a utility function that outputs a single real-valued number.\footnote{Multiattribute utility functions still have a single performance metric for optimization, unlike multi-objective utility functions, which we discuss in Section~\ref{sec:multiobj}.} %

\vparagraph{Utility function depends on a large and context-dependent set of stakeholders} Second, the utility function $G$ should conform to stakeholders' interests. In AD, these stakeholders might include users such as passengers; end consumers%
; partnering businesses like taxi and ride-sharing companies; automotive manufacturers; governmental regulators; providers of research funding; nearby pedestrians, passengers, bicyclists, and residents; and broader society. These stakeholders and the weight given to their interests will differ among vehicles (e.g. with different manufacturers) and will even differ for the same car in different contexts. One such context that could affect $G$ is the driving region, with which values, preferences, and driving culture may differ substantially. Therefore \emph{ideal} reward design for AD might require designing numerous reward functions (or $G$s, more generally). Alternatively, the aforementioned stakeholder-based context could be part of the observation signal, permitting a single monolithic reward function. Such a reward function would allow a policy to be learned across numerous stakeholder-based contexts and generalize to new such contexts. 

\vparagraph{Lack of rigorous methods for evaluating a utility function} Third, when choosing a utility function for algorithmic optimization in some context(s), a critical question arises: given a set of stakeholders, how can one utility function be deemed better or worse than another, and by how much? We have not yet found research on how to measure the degree of a utility function's conformity to stakeholders' interests. 
We also have not found formal documentation of current common practice for evaluating a utility function. Anecdotally, such evaluation often involves both subjectively judging policies that were trained from candidate utility functions and reflecting upon how the utility function might be contributing to undesirable behavior observed in these trained policies. Setting aside the dangers of allowing specific learning algorithms to inform the design of the utility function (see the discussion of trial-and-error reward design in Section~\ref{sec:redflags}), this common approach has not been distilled into a specific process, has not been carefully examined, and in practice varies substantially among different designers of reward functions or other utility functions.
However, our sanity checks in Section~\ref{sec:tools} represent useful steps in this direction.%

\vparagraph{Elicits na{\"i}ve reward shaping} A fourth difficulty may be specific to designing a per-step feedback function like a reward function or a cost function. Driving is a task domain with \emph{delayed} feedback, in that much of the utility of a drive is contained at its end (e.g., based on whether the goal was reached). For drives of minutes or hours, such delayed information about performance renders credit assignment to behavior within a trajectory difficult. In part because of this difficulty, reward shaping appears quite tempting and its na{\"i}ve application is extremely common in research we reviewed (see Section~\ref{sec:shaping}).%

\vsection{Sanity checks for reward functions}
\label{sec:tools}

\begin{table}[]
\begin{spacing}{1}
\begin{footnotesize}
\begin{tabularx}{\textwidth}{ | m{0.18cm} | m{11.3em} | m{17.372em} | m{17.4em} | } 
\hline
  & \textbf{Sanity check failures}         & \textbf{Brief explanation}                                                                                                                                                                 & \textbf{Potential intervention(s)}\\ 
\hline
  
1 & Unsafe reward shaping                            & If reward includes guidance on behavior that deviates from only measuring desired outcomes, reward shaping exists.                                                                                        &  Separately define the true reward function and any shaping reward. Report both true return and shaped return. Change it to an applicable safe reward shaping method. Remove reward shaping. \\ \hline
2 & Mismatch in people's and \newline reward function's preference orderings    & If there is human consensus that one trajectory is better than another, the reward function should agree.                                                                            & Change the reward function to align its preferences with human consensus.                                                                 \\ \hline
3 & Undesired risk tolerance via indifference points & Assess a reward function's risk tolerance via indifference points and compare to a human-derived acceptable risk tolerance.                                                      & Change reward function to align its risk tolerance with human-derived level.                                                         \\ \hline
4 & Learnable loophole(s)                              & If learned policies show a pattern of undesirable behavior, consider whether it is explicitly encouraged by reward.                                                             & Remove encouragement of the loophole(s) from the reward function.                                                                                                     \\ \hline
5 & Missing attribute(s)                               & If desired outcomes are not part of reward function, it is indifferent to them.                                                                                                  & Add missing attribute(s).                                                                                                               \\ \hline
6 & Redundant attribute(s)                             & Two or more reward function attributes include measurements of the same outcome.                                                                                                 & Eliminate redundancy.                                                                                                          \\ \hline
7 & Trial-and-error reward \newline design                    & Tuning the reward function to improve RL agents' performances has unexamined consequences.                                                                                           & Only use observations of behavior to improve the reward function's measurement of task outcomes or to tune separately defined shaping reward.       \\ \hline
8 & Incomplete description of \newline problem specification  & Missing descriptions of reward function, termination conditions, discount factor, or time step duration may indicate insufficient consideration of the problem specification. & In research publications, write the full problem specification and why it was chosen. The process might reveal issues.               \\ \hline
\end{tabularx}
\vspace{1mm}
\caption{Sanity checks one can perform to ensure a reward function does not suffer from certain common problems. Each sanity check is described by what problematic characteristic to look for. Failure of any of the first 4 sanity checks identifies problems with the reward function; failure of the last 4 checks should be considered a warning.}
\label{table:tools}
\end{footnotesize}
\end{spacing}
\end{table}

In this section we develop a set of 8 conceptually simple sanity checks for critiquing and improving reward functions (or cost functions, equivalently). Table~\ref{table:tools} summarizes these sanity checks. Many of the tests apply more broadly to any (trajectory-level) utility function. We demonstrate their usage through critically reviewing reward functions used in RL for AD research. 

The 19 publications we review include every publication on RL for autonomous driving we found that had been published by the beginning of this survey process at a top-tier conference or journal focusing on robotics or machine learning~\cite{dosovitskiy2017carla, isele2018navigating,liang2018cirl,jaritz2018end,cai2019lets,wang2020learning,huegle2019dynamic,toromanoff2020end,paxton2017combining,li2019urban,liu2017learning,kendall2019learning,henaff2019model}, as well as some publications from respected venues that focus more on autonomous driving~\cite{min2019deep,chen2019model,wang2019quadratic,mirchevska2018high,tang2019towards,aradi2018policy}. Of the 19 publications, we arbitrarily designated 10 as ``focus papers'', for which we strove to exhaustively characterize the reward function and related aspects of the task description, typically through detailed correspondence with the authors (see Section~\ref{sec:redflags}). These 10 focus papers are detailed in \ref{app:rewfcns} and~\ref{app:trajcalcs}.

We present 8 sanity checks below, 4 each in their own detailed subsections and 4 in the final subsection, Section~\ref{sec:redflags}. These tests have overlap regarding what problems they can uncover, but each sanity check entails a distinct inquiry. Their application to these 19 publications reveals multiple prevalent patterns of problematic reward design.

\vsubsection{Identifying unsafe reward shaping}
\label{sec:shaping}

In the standard text on artificial intelligence, Russell and Norvig assert, ``As a general rule, it is better to design performance metrics according to what one actually wants to be achieved in the environment, rather than according to how one thinks the agent should behave''~\citep[p.~39]{russell2020artificial}. In the standard text on RL, \citet[p.~54]{sutton2018reinforcement} agree in almost the same phrasing, adding that imparting knowledge about effective behavior is better done via the initial policy or initial value function.
More succinctly, \emph{specify how to measure outcomes, not how to achieve them}. %
Exceptions to this rule should be thoughtfully justified.

Yet using rewards to encourage and hint at generally desirable behavior---often with the intention of making learning more efficient and tractable when informative reward is infrequent or inaccessible by most policies---is intuitively appealing. This practice has been formalized as reward shaping, in which the learning agent's received reward is the sum of true reward %
and shaping reward. The boundary between these two types of rewards is not always clear when the ``true'' objective is not given, such as in AD. Nonetheless, some rewards are more clearly one of the two types. 

The dangers of reward shaping are well documented~\cite{randlov1998learning,ng1999piu,aamas12-knox}. These dangers include creating ``optimal'' policies that perform catastrophically. Perhaps worse, reward shaping can appear to help by increasing learning speed without the reward designer realizing that they have, roughly speaking, decreased the upper bound on performance by changing the reward function's preference ordering over policies. %

There is a small canon of reward-shaping research that focuses on how to perform reward shaping with certain \emph{safety} guarantees~\cite{ng1999piu,wiewiora2003potential,asmuth2008potential,devlin2012dynamic,grzes2017reward}. \emph{Safety} here means that the reward shaping has some guarantee that it will not harm learning, which differs from its colloquial definition of avoiding harm to people or property. A common safety guarantee is \emph{policy invariance}, which is having the same set of optimal policies with or without the shaping rewards. We generally recommend that attempts to shape rewards be informed by this literature on safe reward shaping. For all techniques with such guarantees, shaping rewards are designed \emph{separately} from the true reward function. Also, if possible, the utility function $G$ that arises from the true reward should be equivalent to the main performance metric used for evaluating learned policies.%

We now formulate the above exposition as a sanity check. Unsafe reward shaping can be identified by first identifying reward shaping---without regard to safety---and then determining whether the designers of the reward function are either following a known safe reward shaping method or have a persuasive argument that their shaped rewards are safe.

\vparagraph{Application to AD} Acknowledging the subjectivity of classifying whether reward is shaped when shaping is not explicitly discussed, we confidently judge that, of the 19 publications we surveyed, 13 included reward shaping via one or more attributes of their reward functions~\cite{dosovitskiy2017carla,liang2018cirl,jaritz2018end,min2019deep,wang2020learning,chen2019model,huegle2019dynamic,toromanoff2020end,paxton2017combining,li2019urban,liu2017learning,tang2019towards,henaff2019model}. Another 2 included reward attributes which could arguably be considered reward shaping~\cite{wang2019continuous,aradi2018policy}.
Examples of behavior encouraged by reward shaping in these 13 publications are staying close to the center of the lane~\citep{jaritz2018end}, passing other vehicles~\citep{min2019deep}, not changing lanes~\citep{huegle2019dynamic}, increasing distances from other vehicles~\citep{wang2020learning},
avoiding overlap with the opposite-direction lane~\citep{dosovitskiy2017carla,liang2018cirl}, and steering straight at all times~\citep{chen2019model}. Other examples can be found in \ref{app:shaping}. %
All of these encouraged behaviors are heuristics for \emph{how to achieve good driving}---violating the aforementioned advice of Russell, Norvig, Sutton, and Barto---and it is often easy to construct scenarios in which they \emph{discourage} good driving. For example, the reward shaping attribute that penalizes changing lanes~\citep{huegle2019dynamic} would discourage moving to lanes farther from other vehicles or pedestrians, including those acting unpredictably.
Many of the examples above of behaviors encouraged by shaping rewards might be viewed as metrics that are not attributes of the true utility function yet are highly and positively \emph{correlated} with performance. As \citet{amodei2016concrete} discussed, rewarding behavior that is correlated with performance can backfire, since strongly optimizing such reward can result in policies that trade increased accumulation of the shaping rewards for large reductions in other performance-related outcomes, driving down the overall performance. This concept has been memorably aphorized as Goodhart's law: ``When a [proxy] measure becomes a target, it ceases to be a good measure.'' \cite{strathern1997improving,goodhart1984problems}.

Ostensibly, a similar criticism could be aimed at measures that \textit{should} be part of the true reward function. For instance, assume that reducing gas cost and avoiding collisions are attributes of the true utility function. Then reducing gas cost could discourage accelerating, even when doing so would avoid a potential collision. The critical difference, however, between attributes measuring desired outcomes and reward-shaping attributes is that trading off two or more true attributes of the utility function can result in higher overall utility and therefore be desirable. In our example above, reducing a \textit{correctly weighted} gas cost would presumably have negligible effect on the frequency of collisions, since the benefit of avoiding collisions would far outweigh the benefit of reducing gas cost. Another perspective on this difference is that effective optimization can increase \textit{reward-shaping attributes} at the expense of overall utility, whereas effective optimization may increase \textit{true utility attributes} at the expense of other utility attributes, but not at the expense of overall utility.

Of the 13 publications which use reward functions we are confident are shaped, 8 were in the set of 10 focus papers~\cite{dosovitskiy2017carla,liang2018cirl,jaritz2018end,min2019deep,wang2020learning,chen2019model,huegle2019dynamic,toromanoff2020end}. None of the 8 papers explicitly described the separation between their shaping rewards from their true rewards, and none discussed policy invariance or other guarantees regarding the safety of their reward shaping. 
Of these 8 papers, only \citeauthor{jaritz2018end} and  \citeauthor{toromanoff2020end} acknowledged their usage of reward shaping, and only the former discussed its undesirable consequences. %
\citeauthor{jaritz2018end} write ``the bots do not achieve optimal trajectories ...  [in part because] the car will always try to remain in the track center'', which their reward function explicitly incentivizes.
Further, in most of these 8 papers with reward shaping, the performance of learned policies was not compared in terms of their return but rather according to one or more other performance metrics, obscuring how much the undesirable behavior (e.g., frequent collisions) was a result of the RL algorithm's imperfect optimization or of the reward function it was optimizing against. In their work on learning reward functions, \citet{ibarz2018reward} provide other useful examples of how to analyze an alternative reward function against returns from the true reward function. 3 of these 8 papers did report return~\cite{wang2020learning,huegle2019dynamic,chen2019model}. %

\vsubsection{Comparing preference orderings}
\label{sec:preforder}

Although it is difficult for humans to score trajectories or policies in ways that are consistent with utility theory, simply judging one trajectory as better than another is sometimes easy. Accordingly, one method for critiquing a utility function is to compare the utility function's trajectory preferences to some ground-truth preferences, whether expressed by a single human or a decision system among multiple stakeholders, such the issuance of regulations.
This preference comparison test of a utility function $G$ is $\tau_A \prec \tau_B \iff G(\tau_A) < G(\tau_B)$, where $\prec$ means ``is less preferred than''. Finding a $\tau_A$ and $\tau_B$ for which this statement is false indicates a flaw in the utility function but does not by itself evaluate the severity of the flaw. However, severity is implied when one trajectory is strongly preferred. Note that we focus here on \textit{evaluating} a utility function, which differs from \textit{learning} a utility function from preferences over trajectories or subtrajectories (see Section~\ref{sec:rewlearning}), after which this sanity check and others can be applied.

\begin{wrapfigure}{}{0.27\textwidth}
    \vspace{-5mm}
    \includegraphics[width=1\linewidth]{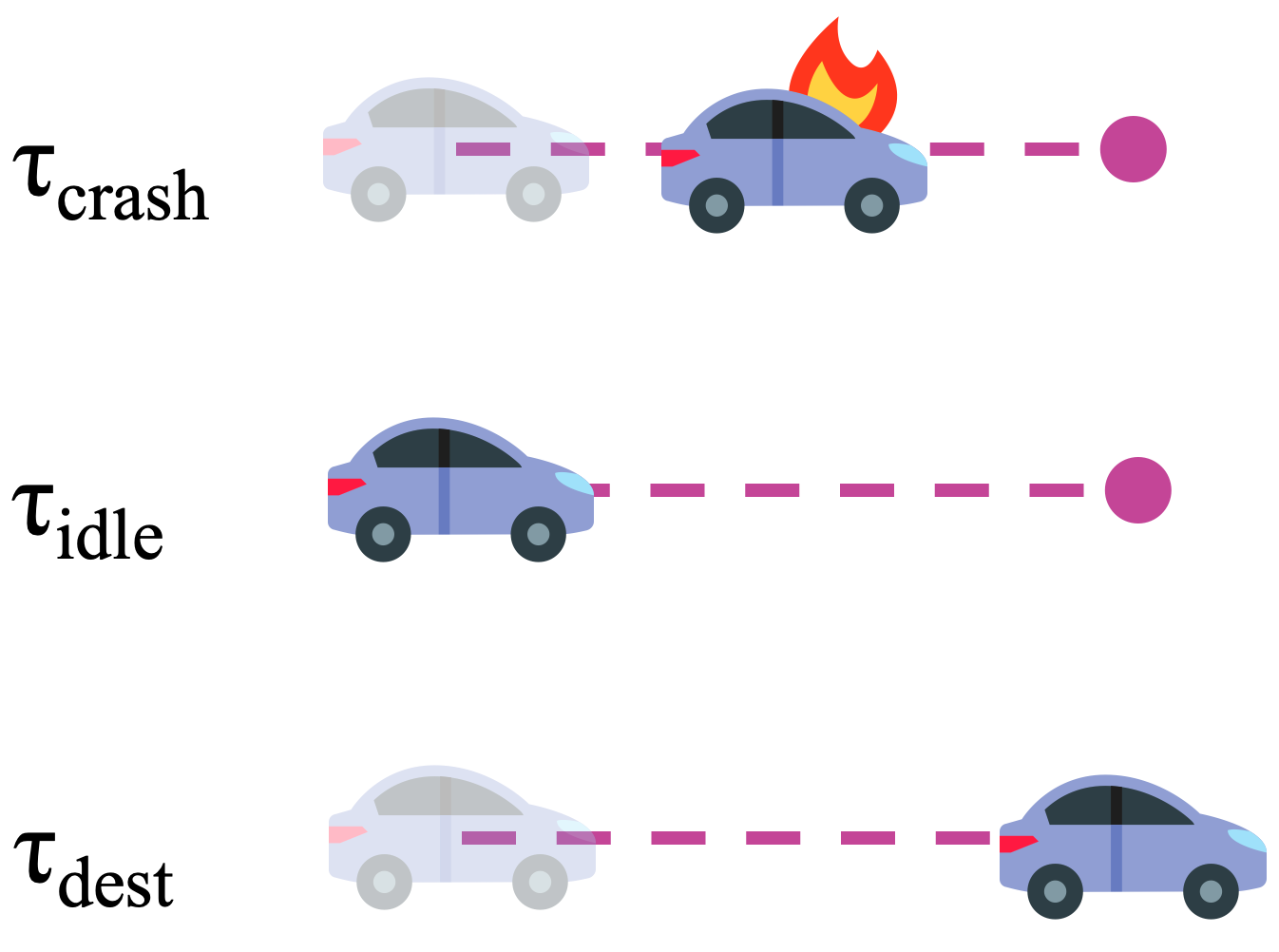}
    \vspace{-2mm}
    \caption{\small{Illustrations of the abstract trajectories used in Sections~\ref{sec:preforder} and \ref{sec:indifference}}.}
    \vspace{-8mm}
    \label{fig:trajectories}
\end{wrapfigure}

\vparagraph{Application to AD}
We apply this comparison by choosing two trajectories such that, under non-exceptional circumstances, one trajectory is strongly preferred. We specifically let $\tau_{crash}$ be a drive that is successful until crashing halfway to its destination and let $\tau_{idle}$ be the safe trajectory of a vehicle choosing to stay motionless where it was last parked. Figure~\ref{fig:trajectories} illustrates $\tau_{crash}$ and $\tau_{idle}$. %
Of the 10 focus papers, 9 permit estimating $G(\tau_{crash})$ and $G(\tau_{idle})$. \citet{huegle2019dynamic} does not (see \ref{app:trajcalcs}).
For the calculation of these utilities here and later in this article, we assume reward is not temporally discounted in the \emph{problem specification}---which is generally considered correct for episodic tasks~\cite[p.~68]{sutton2018reinforcement} like these---despite nearly all papers' adherence to the current best practice of discounting future reward to aid deep reinforcement learning \emph{solutions} (as discussed by \citet{pohlen2018observe}).%

\begin{wrapfigure}{}{0.65\textwidth}
    \centering
    \includegraphics[width=.95\linewidth]{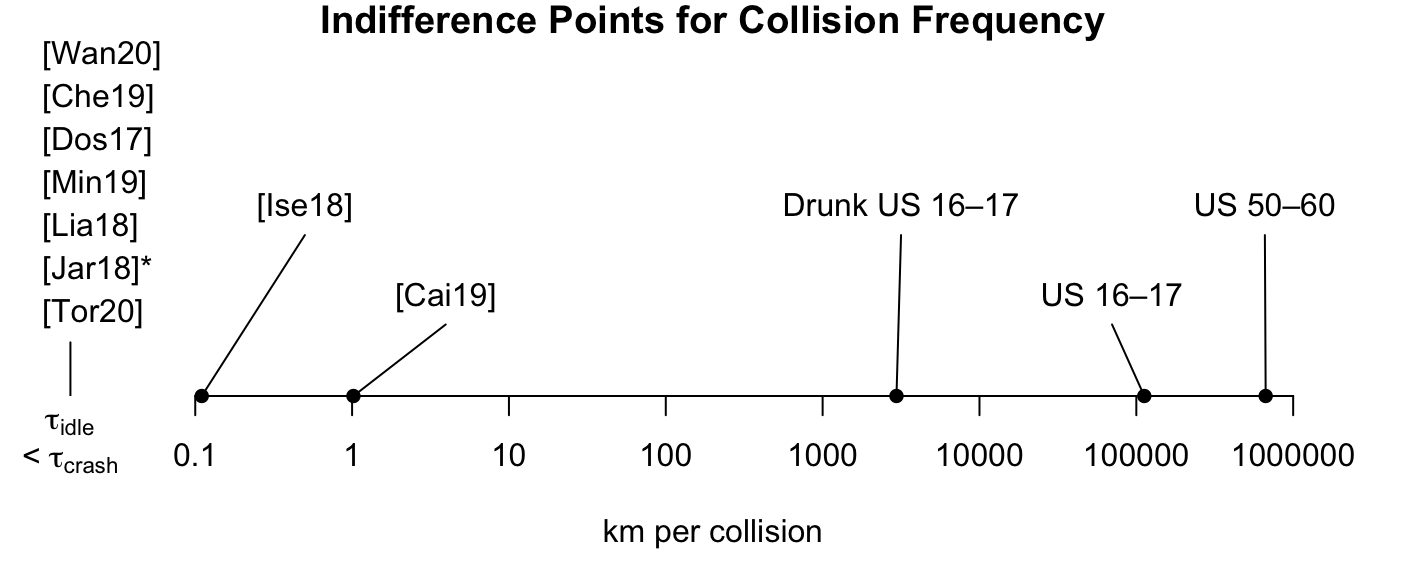}
    \vspace{-2mm}
    \caption{\small{Estimates of kilometers per collision at which various published reward functions are indifferent regarding whether they prefer safely declining to move or driving with a certain km per collision rate. Higher values indicate stronger safety requirements. Publications are referenced by the first 3 letters of their first author's name and the last two digits of their publication year. \emph{Che19} refers to the publication by Jianyu Chen. The 3 points on the right designate estimates of actual km per collision for the age group among US drivers with the most km per collisions (50--60 year olds) and the least (16--17 year olds)~\cite{tefft2017rates}, as well as a rough estimate of km per collision for a drunk 16--17 year old (from applying a 37x risk for blood alcohol concentration $\geq 0.08$, as estimated by ~\citet{peck2007improved}). %
    {\fontfamily{phv}\selectfont *}The task domain of \citet{jaritz2018end} was presented as a racing video game and therefore should not be judged by real-world safety standards.}}
    \label{fig:crashrates}
\end{wrapfigure}

We presume that any appropriate set of stakeholders would prefer a vehicle to be left idle rather than to proceed to a certain collision: $\tau_{crash} \prec \tau_{idle}$. Yet of these 9 evaluated reward functions, only 2 have reward functions with the correct preference and 7 have reward functions that would prefer $\tau_{crash}$ and its collision. These 7 papers are identified on the left side of Figure~\ref{fig:crashrates}, under $\tau_{idle} \prec \tau_{crash}$. We do not calculate utilities for the reward functions of the 9 papers that were \emph{not} in the set of focus papers, but our examination of these other reward functions suggests a similar proportion of them would likewise have an incorrect ordering.

Calculation of returns for these trajectories allows a much-needed sanity check for researchers conducting RL-for-AD projects, avoiding reward functions that are egregiously dangerous in this particular manner.

\vsubsection{Comparing indifference points}
\label{sec:indifference}

A more complex form of preference comparison reveals problems in the 2 papers that passed the test in Section~\ref{sec:preforder}. For three trajectories $\tau_A \prec \tau_B \prec \tau_C$, the continuity axiom of utility theory states that there is some probability $p$ such that a rational agent is indifferent between (1) $\tau_B$ and (2) sampling from a Bernoulli distribution over $\tau_A$ and $\tau_C$, where $\tau_C$ occurs with probability $p$~\cite{von1944theory}: 
\begin{align}
    G(\tau_B) &= p G(\tau_C) + (1-p)G(\tau_A).
\end{align} 
This indifference point $p$ can often be compared to a ground-truth indifference point derived from human stakeholders that reveals their risk tolerance.

The use of indifference points for AD below is an exemplar of a widely applicable methodology for testing a reward function. This methodology entails choosing $\tau_B$ to be a trajectory that can be achieved with certainty, often by some default behavior or inaction. $\tau_A$ and $\tau_C$ are then chosen to contain two possible outcomes from a risky departure from default behavior or inaction, where $\tau_C$ is a successful risky outcome and $\tau_A$ is an unsuccessful risky outcome. From another perspective, $\tau_A$ is losing a gamble, $\tau_B$ is not gambling, and $\tau_C$ is winning a gamble. One set of examples for $(\tau_A, \tau_B, \tau_C)$ in a domain other than AD includes trajectories from when a video game agent can deterministically finish a level with only one action: (seeking more points but dying because time runs out, finishing the level without seeking additional points, getting more points and finishing the level before time runs out).

\vparagraph{Application to AD}
To apply this test of preferences over probabilistic outcomes to AD, we add $\tau_{succ}$ to $\tau_{crash}$ and $\tau_{idle}$ from Section~\ref{sec:preforder}, where $\tau_{succ}$ is a trajectory that successfully reaches the destination. $\tau_{crash} \prec \tau_{idle} \prec \tau_{succ}$. Therefore, choosing $p$ amounts to setting permissible risk of crashing amongst otherwise successful trips. In other words, a policy that has higher risk than this threshold $p$ is less preferred than one that refuses to drive at all. Human drivers appear to conduct similar analyses, sometimes refusing to drive when faced with a significant probability of collision, such as during severe weather. Figure~\ref{fig:crashrates} displays the calculated $p$ converted to a more interpretable metric: km per collision\footnote{$\text{Kilometers per crash at indifference point} = ([p/(1-p)]+0.5) \times \text{distance of a  successful path}$, where $p/(1-p)$ is the amount of $\tau_{succ}$ trajectories per half-length $\tau_{crash}$ trajectory at the indifference point.} at which driving is equally preferable to not deploying a vehicle. For comparison, we also plot estimates of police-reported collisions per km for various categories of humans. These human-derived indifference points provide very rough bounds on US society's indifference point, since drunk driving is considered illegal and 16--17 year old US citizens are permitted to drive. As the figure shows, of those $9$ focus papers that permit this form of analysis, $0$ require driving more safely than a legally drunk US 16--17 year old teenager. %
The most risk-averse reward function by this metric~\cite{cai2019lets} would approve driving by a policy that crashes 2000 times as often as our estimate of drunk 16--17 year old US drivers.%

An argument against this test---and more broadly against requiring the utility function to enforce driving at or above human-level safety---is that penalizing collisions too much could cause an RL algorithm to correctly learn that its current policy is not safe enough for driving, causing it to get stuck in a conservative local optimum of not moving. This issue however can potentially be overcome by creating sufficiently good starting policies or by performing reward shaping explicitly and rigorously. Further, there is a significant issue with the argument above, the argument that the reward function should encourage the RL algorithm to gather driving experience by being extremely lenient with collisions. In particular, whether a specific weighting of a collision penalty will effectively discourage collisions without making the vehicle avoid driving is dependent on the performance of the current policy. As the RL algorithm improves its driving, the \textit{effective} collision-weighting values would generally need to increase. The true reward function is part of a task specification and should not change as a function of the policy. However, such dynamic weighting could be achieved by defining the true reward function to have a risk tolerance that is desirable in real-world situations then adjusting the weight given to collisions via a form of dynamic reward shaping: the collisions weight would start small and be gradually increased to scaffold learning while the policy improves, eventually reaching its true weight value.  In this strategy, reward shaping is temporary and therefore policy invariance is achieved once shaping ends.

\begin{wrapfigure}{}{0.32\textwidth}
    \vspace{-13mm}
    \includegraphics[width=1\linewidth]{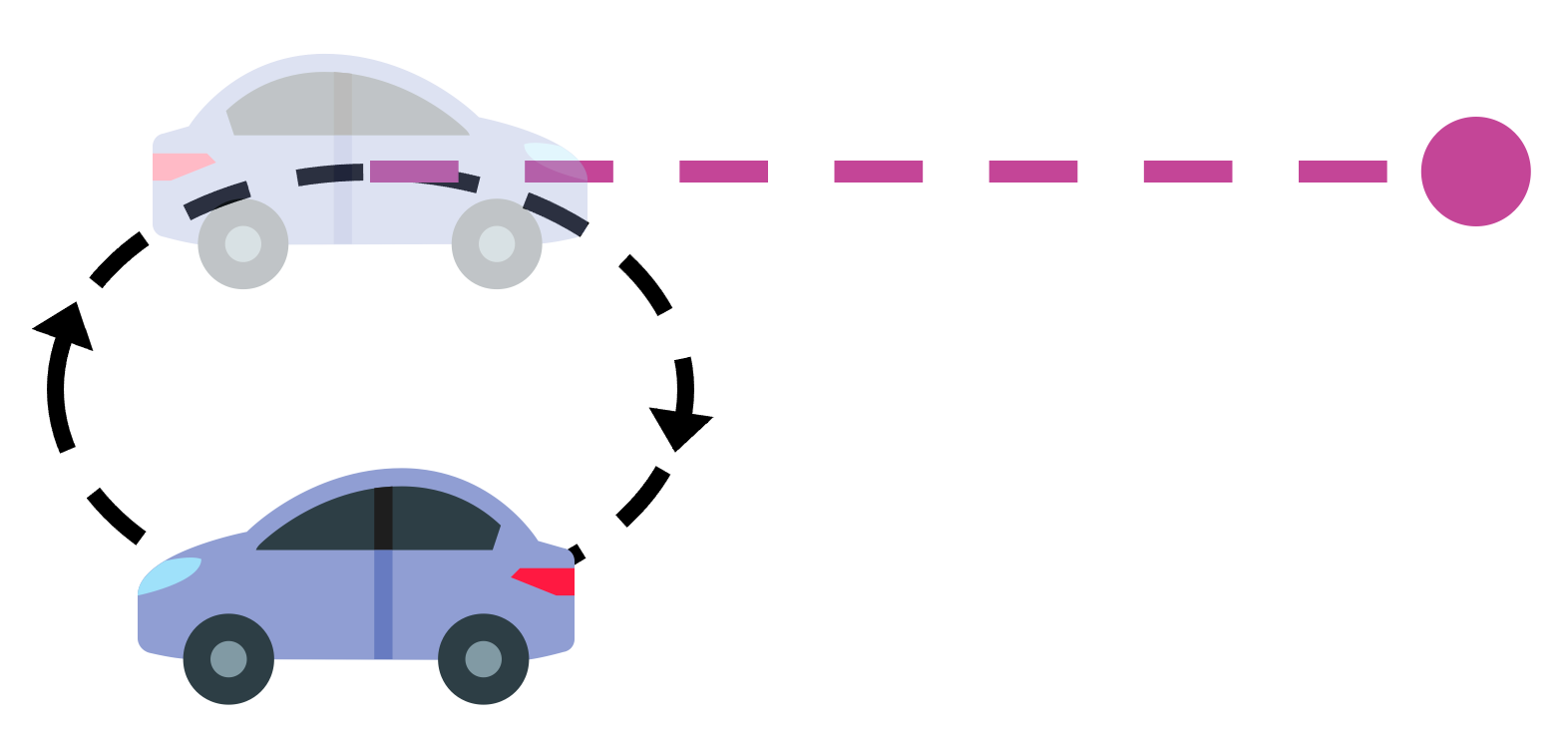}
    \vspace{-4mm}
    \caption{\small{Illustration of a common learnable loophole, in which the agent moves in circles to repeatedly collect reward for progress, never reaching the goal.}}
    \vspace{-3mm}
    \label{fig:loophole}
\end{wrapfigure}

\vsubsection{Identifying learnable loopholes}

Once a designed reward function is used for learning a policy, observable patterns of undesirable behavior might emerge. When such behavior increases utility, it is often referred to as \emph{reward hacking} or \emph{specification gaming} (terms that implicitly and unfairly blame the agent for correctly optimizing a flawed utility function). %
Colloquially, such technically legal violations of the spirit of the law are called \emph{loopholes}.
The RL literature provides numerous catastrophic examples of such loopholes~\cite{randlov1998learning,amodei2016faulty,ng1999piu}, which often involve learned trajectories that actually loop in physical space to repeatedly accrue some reward but prevent long-term progress (see Figure~\ref{fig:loophole}).\footnote{More examples can be found at this webpage: \href{https://docs.google.com/spreadsheets/u/1/d/e/2PACX-1vRPiprOaC3HsCf5Tuum8bRfzYUiKLRqJmbOoC-32JorNdfyTiRRsR7Ea5eWtvsWzuxo8bjOxCG84dAg/pubhtml}{``Specification gaming examples in AI''}}
However, these loopholes can also be subtle, not dominating performance but nonetheless limiting it, and in such cases loopholes are more difficult to find through observations of learned behavior. In general, both subtle and blatant loopholes might be found by observing undesirable behavior and reflecting that the reward function encourages that behavior. A more rigorous improvement on such reflection is to estimate the utility (or return, equivalently) for an observed trajectory that contains the undesirable behavior and also estimate the utility for that trajectory modified minimally to not contain the undesirable behavior; if the less desirable trajectory receives more utility, then a loophole likely exists. An alternative perspective on this improvement is that it is a method for finding two trajectories with which to perform the sanity check of comparing preference orderings (Section~\ref{sec:preforder}).  %

\vparagraph{Application to AD}
We do not observe blatantly catastrophic loopholes in the RL-for-AD literature, perhaps because most reward functions for AD thus far appear to have been designed via trial and error (see Section~\ref{sec:redflags}); in such trial-and-error design, any such catastrophic loopholes would likely get caught, and the reward function would then be tuned to avoid them. However, the learned-policy limitation that~\citet{jaritz2018end} discussed (see Section~\ref{sec:shaping}) is an example of a learned loophole in RL for AD.%

\vsubsection{Sanity checks for which failure is a warning} 
\label{sec:redflags}

In addition to the sanity checks described above, other methods can be used to raise red flags, indicating \textit{potential} issues in the reward function that warrant further investigation. Because the descriptions of these sanity checks are relatively short, the discussion of their applications to AD are less explicitly separated than for previously described sanity checks.

\vparagraph{Missing attributes}
Utility functions sometimes lack attributes needed to holistically judge the performance of a trajectory. Specifically, as discussed in \citet{amodei2016concrete}, omission of an attribute ``implicitly expresses indifference'' regarding that attribute's measurements of outcomes. A simple example in autonomous driving is to ignore passenger experience while including an attribute that increases utility for making progress towards a destination; the learned behavior may exhibit high acceleration and jerk, which tend to harm passenger experience but do not directly affect this example utility function.
Section~\ref{sec:challenge} contains a list of potential attributes for autonomous driving that may help in this evaluation. %
Identifying missing attributes and adding them is one solution. When inclusion of all missing attributes is intractable, Amodei et al. propose penalizing a policy's impact on the environment. However, this approach, called \emph{impact regularization}, shares potential issues with reward shaping.%

\vparagraph{Redundant attributes}
Similarly, a utility function may have redundant attributes that encourage or discourage the same outcomes. Such overlap can overly encourage only part of what constitutes desirable outcomes. The overlap can also complicate a reward designer's understanding of the attributes' combined impact on the utility function. For instance, consider an autonomous driving utility function that includes two attributes, one that penalizes collisions and one that penalizes repair costs to the ego vehicle. Both attributes penalize damage to the ego vehicle from a collision. Yet each attribute also includes a measurement of outcomes that the other does not: a collision penalty can discourage harm to people and external objects, and a repair-costs penalty discourages driving styles that increase the rate of wear. When redundant attributes exist in the utility function, solutions include separating redundant aspects of multiple attributes into a new attribute or removing redundant aspects from all but one attribute. Executing these solutions is not always straightforward, however. In our example above, perhaps the \emph{collision} penalty could be separated into components that measure harm to humans and animals, harm to external objects, and increased repair costs for the ego vehicle's maintainer. Then the repair-costs component of the collision penalty could be removed, since it is already contained within the penalty for overall repair costs.

\vparagraph{Trial-and-error reward design}
If a publication presents its reward function without describing the reward design process, we suspect that it was likely designed through a process of trial and error. 

This \emph{trial-and-error reward design} process involves designing a reward function, testing an RL agent with it, \emph{using observations of the agent's learning to tune the reward function}, and then repeating this testing and tuning process until satisfied. This process is also described by \citet[p.~469]{sutton2018reinforcement}.
Since the reward function itself is being revised, typically one or more other performance metrics are employed to evaluate learned policies. Those performance metrics could be based on subjective judgment or be explicitly defined.

One issue with this trial-and-error reward design is that the specification of the reinforcement learning problem should not in principle be adjusted to benefit a candidate solution to the problem. 
More practically, another issue is that this manual reward-optimization process might overfit. In other words, trial-and-error reward design might improve the reward function's efficacy---whether measured by the reward designer's subjective evaluation, another performance metric, or otherwise---in the specific context in which it is being tested, but then the resultant reward function is used in other untested contexts, where its effects are unknown. Factors affecting trial-and-error reward design include both the RL algorithm and the duration of training before assessing the learned policy. In particular, after the designer chooses a final reward function, we suspect they will typically allow the agent to train for a longer duration than it did with the multiple reward functions evaluated during trial-and-error design. Further, any comparison of multiple RL algorithms will tend to unfairly favor the algorithm used during trial-and-error design, since the reward function was specifically tuned to improve the performance of that RL algorithm.

Two types of trial-and-error reward design do appear appropriate. The first type is when intentionally designing a reward shaping function, since reward shaping is part of an RL solution and is therefore not changing the RL problem. %
The second type is when observations of learned behavior change the designer's understanding of what the task's trajectory-level performance metric should be, and the reward function is changed \textit{only} to align with this new understanding.

Trial-and-error reward design
for AD is widespread: of the 8 publications whose authors shared their reward design process with us in correspondence, all 8 reported following some version of defining a linear reward function and then manually tuning the weights or revising the attributes via trial and error until the RL algorithm learns a satisfying policy~\cite{dosovitskiy2017carla,isele2018navigating,cai2019lets,huegle2019dynamic,min2019deep,chen2019model,wang2020learning,toromanoff2020end}.%
Based on informal conversations with numerous researchers, we suspect trial-and-error reward design is widespread across many other domains too. Yet we are unaware of any research that has examined the consequences of this ad hoc reward-optimization process.

\vparagraph{Incomplete problem specification in research presentations}
Many publications involving RL do not exactly specify aspects of their RL problem specification(s). 
Only 1 of the 10 focus papers thoroughly described 
the reward function, discount factor, termination conditions and time step duration used~\citet{liang2018cirl}. We learned the other 9 papers' specifications through correspondence with their authors.
We conjecture that a broader analysis of published reward functions would find that such omission of problem specification details is positively correlated with reward design issues. 
In the absence of such confirmatory analysis, we nonetheless encourage authors to write the full problem specification, both for the reader's benefit and because the practice of writing the problem details may provide insights.

\vsection{Exploring the design of an effective reward function}
\label{sec:design}

In contrast to the previous section---which focuses on how \textit{not} to design a reward function---this section instead considers \textit{how} to design one. This section contains exploration that we intend to be preliminary to a full recommendation of a specific reward function or of a process for creating one. Again, AD serves as our running example throughout this section. 

\vsubsection{Performance metrics beyond RL}

One potential source of reward-design inspiration is the performance metrics that have been created by communities beyond RL researchers.

For AD specifically, prominent among these metrics are those used by regulatory agencies and companies developing AD technology. Many such metrics express the \textit{distance per undesirable event}, which incorporates two critical utility-function attributes: making progress and avoiding failure states like collisions. Specifically, the California Department of Motor Vehicles (DMV) requires reporting of \textit{miles per disengagement}, where a disengagement is defined as the deactivation of the vehicle's autonomous mode and/or a safety driver taking control from the autonomous system. Criticisms of this metric include that it ignores important context, such as the complexity of driving scenarios, the software release(s) being tested, and the severity of the outcomes averted by the safety drivers' interventions. The California DMV also requires a report to be filed for each collision, through which a \textit{miles per collision} measure is sometimes calculated~\cite{favaro2017examining}.
This metric is vulnerable to some of the same criticisms. Also, note that disengagements often prevent collisions, so safety drivers' disengagement preferences can decrease miles per disengagement while increasing miles per collision, or vice versa; therefore, the two complementary metrics can be combined for a somewhat clearer understanding of safety. Another metric is miles per fatality, which addresses the ambiguity of a collision's severity. 
In contrast to the 8 attributes we listed in Section~\ref{sec:challenge} as important for an AD utility function, each of the above metrics only covers 2 attributes---distance traveled and a count of an undesirable event.

Performance metrics for AD have also been designed by other robotics and artificial intelligence communities. In particular, per-time-step cost functions developed by the planning and control communities could be converted to reward functions by multiplying their outputs by $-1$. Additionally, insight might be gained by examining the reward functions learned by techniques reviewed in Section~\ref{sec:rewlearning}. However, a review of such cost functions and learned reward functions is beyond the scope of this article.%

\vsubsection{An exercise in designing utility function attributes}
\label{sec:exercise}

To get a sense of the challenges involved in expressing reward function attributes, let us consider in detail three attributes that we listed previously in Section~\ref{sec:challenge}: progress to the destination, obeying the law, and passenger experience. We assume that each attribute is a component in a linear reward function, following the common practice. For simplicity, we focus on the scenario of driving one or more passengers, without consideration of other cargo.

We see one obvious candidate for \textbf{progress to the destination}.\footnote{For reasons hinted at in this subsection, we do not focus on the simpler binary attribute of reaching the destination, which ignores that drop-off points other than the destination can have many different effects on passengers' utilities.}
This candidate attribute is the proportion of progress towards the destination, expressed as a value $\leq 1$ and calculated at time step $t$ as
{\small
\begin{align*}
    \frac{(\text{\emph{route length from position at time }}t) - (\text{\emph{route length from position at time }}t-1)}{ \text{\emph{route length from start}}},
\end{align*} 
}
which adds to $1$ over an entire successful trajectory.
(Route length can be calculated in various ways, such as the distance of the shortest legal path that can be planned.) 

This approach initially appears reasonable but nonetheless has at least one significant issue: 
each equal-distance increment of progress has the same impact on the attribute. To illustrate, consider two policies. One policy always stops exactly halfway along the route. The other policy reaches the destination on half of its trips and does not start the trip otherwise. 
With respect to the progress-to-the-destination attribute proposed above, both polices have the same expected performance. Yet the performance of these two policies do not seem equivalent. For our own commutes, we authors certainly would prefer a ride-sharing service that cancels half the time than a service that always drops us off halfway to our destination.

This dilemma leaves open the question of how to calculate progress to the destination as an attribute. Perhaps it should be based upon the utility derived by the passenger(s) for being transported to some location. We suspect this utility would \emph{generally} be lowest when the drop-off point is far both from the pickup location and from the destination. A highly accurate assessment of this utility would seemingly require information about what the passenger(s) would do at any drop-off location, including their destination. For instance, if a drug store is the destination, a passenger's utility at various drop-off locations will differ greatly based on whether they plan to purchase urgently needed medication or to buy a snack. Such information is unlikely to be available to autonomous vehicles, but nonetheless a coarse estimate of a passenger's utility for a specific drop-off location could be made with whatever information that is available. Also, perhaps passengers could opt in to share that reaching their destination is highly consequential for them, allowing the driving policy to reflect this important information.

\textbf{Obeying the law} is perhaps even trickier to define precisely as an attribute. This attribute could be expressed as the penalties incurred for breaking laws. Unfortunately, such penalties come in different units, with no obvious conversion to the other units---such as fines, time lost interacting with law enforcement and court systems, and maybe even time spent incarcerated---making difficult the combination of these penalties into a single numeric attribute. An additional issue is that modeling the penalties ignores the cost to others, including to law enforcement systems. If such \emph{external costs} are included, the attribute designer might also want to be careful that some of those external costs are not redundantly expressed in another attribute, such as a collisions attribute. A further challenge is obtaining a region-by-region encoding of driving laws and their penalties.

Passengers are critical stakeholders in a driving trajectory, and \textbf{passenger experience} appears important because their experiences are not always captured by other metrics like collisions. For example, many people have experienced fear as passengers when their driver brakes later than they prefer, creating a pre-braking moment of uncertainty regarding whether the driver is aware of the need to slow the vehicle, which they are. Though the situation was actually safe in hindsight%
, the late application of brakes created an unpleasant experience for the passenger. %
To define passenger experience as an attribute, one candidate solution is to calculate a passenger-experience score through surveys of passenger satisfaction, perhaps by averaging all passengers' numeric ratings at the end of the ride.
However, surveys rely on biased self report and might be too disruptive to deploy for every trajectory. In experimental design, when surveys are asking for respondents' predictions of their own behavior, it is advisable to consider whether instead the experiment could rely on observations of behavior; this guideline prompts the question of whether future passenger behavior---such as choosing to use the same AD service again---might be more useful as part of the attribute than direct surveys. %
Lastly, passenger experience is an underdefined concept, despite our confidence that it is important to include in the overall AD utility function, leaving open the question of what exactly to measure.

\textbf {Combining these attributes} is also difficult, since the units of each attribute might not be straightforwardly convertible to those of other attributes. The reward function is commonly a linear combination of attributes; however, this linearity assumption could be incorrect, for example if the utility function needs to be the result of a conjunction over attributes, such as a binary utility function for which success is defined as reaching the destination without collision. Also, weight assignment for such linearly expressed reward functions is often done by trial and error (see Section~\ref{sec:redflags}), possibly in part because the researchers lack a principled way to weigh attributes with different units. %

\vsubsection{A financial utility function for AD}
One potential solution to the challenge of how to combine attributes is to express all attributes in the same unit, so they can be added without weights. Specifically, a financial utility function might output the change in the expectation of net profit across all stakeholders caused by $\tau$. 
 Utilities expressed in currency units are common in RL when profit or cost reduction are the explicit goals of the task, such as stock trading~\cite{nevmyvaka2006reinforcement} and tax collection~\cite{miller2012tax}, but we are unaware of its usage as an optimization objective for AD. %

To create such a financial utility function for AD, non-financial outcomes would need to be mapped to financial values, perhaps via an assessment of people's willingness to pay for those outcomes.
We have been surprised to find that some non-financial outcomes of driving have a more straightforward financial expression than we initially expected, providing optimism for this strategy of reward design. For example, much effort has gone towards establishing a \textit{value of statistical life}, which allows calculation of a monetary value for a reduction in a small risk of fatalities. The value of statistical life is used by numerous governmental agencies to make decisions that involve both financial costs and risks of fatality. The US Department of Transportation's value of statistical life was \$11.6 million US Dollars in 2020~\cite{timothy2021usdot-vsl,timothy2021usdot-vsl-treatment}.

\vsubsection{Methods for learning a reward function}
\label{sec:rewlearning}

Instead of manually designing a reward function, one can instead learn one from various types of data. Methods of learning reward functions include inverse reinforcement learning from demonstrations~\cite{ng2000algorithms,ziebart2008maximum}, learning reward functions from pairwise preferences over trajectories~\cite{wirth2017survey,christiano2017deep,brown2019extrapolating}, and inverse reward design from trial-and-error reward design in multiple instances of a task domain~\cite{hadfield2017inverse}. %
Traditionally, these approaches assume that reward is a linear function over pre-specified attributes and that only the weights are being learned, so the challenges of choosing attributes remain. Approaches that instead model reward via expressive representations like deep neural networks~\cite{wulfmeier2015maximum,fu2017learning} could avoid this challenge. Another issue is that there is no apparent way to evaluate the result of reward learning without already knowing the utility function $G$, which is not known for many tasks like autonomous driving; blindly trusting the results of learning is particularly unacceptable in safety-critical applications like AD. For these methods, the sanity checks we present in Section~\ref{sec:tools} could provide partial evaluation of such learned reward functions.

\vsubsection{Multi-objective reinforcement learning}
\label{sec:multiobj}

Multi-objective approaches~\cite{roijers2013survey} are an alternative to defining a \textit{single} utility or reward function for optimization. In particular, the combination of the attributes of the utility function could be left undefined until after learning. Such an approach may fit well with autonomous driving, for which some attributes change over time. For example, the frequent price changes of petroleum gasoline or electricity could be accounted for by proportionally re-weighting the fuel costs attribute. Many multi-objective reinforcement learning algorithms evaluate sets of policies such that the best policy for a \textit{specific} utility function parametrization can later be estimated without further learning. For linear utility functions---including those that arise from linear reward functions---successor features and generalized policy improvement~\cite{dayan1993improving,barreto2016successor} are promising techniques to make decisions under a changing utility function without requiring further task experience. Much of this article's inquiry and the sanity checks in Section~\ref{sec:tools} apply to the choice of utility-function parametrization that often must be made after multi-objective RL to enable task execution.

\vsection{Conclusion}
\label{sec:conclusion}

In the US alone, 1.9 trillion miles were driven in 2019~\cite{vehiclemiles2020}. Once autonomous vehicles are prevalent, they will generate massive amounts of experiential data. Techniques like reinforcement learning can leverage this data to further optimize autonomous driving, whereas many competing methods cannot do so (e.g., behavioral cloning) or are labor-intensive (e.g., manual design of decision-making). Despite the suitability of RL to AD, it might not play a large role in its development without well-designed objectives to optimize towards. By using AD as a motivating example, this article sheds light on the problematic state of reward design, and it provides arguments and a set of sanity checks that could jump start improvements in reward design. 
We hope this article provokes conversation about reward specification---for autonomous driving and more broadly---and adds momentum towards a much-needed sustained investigation of the topic. 

From this article, we see at least six impactful directions for further work. First, specifically for AD, one could craft a reward function or other utility function for AD that passes our sanity checks, includes attributes that incorporate all relevant outcomes, addresses the issues discussed in Section~\ref{sec:exercise}, and passes any other tests deemed critical for an AD utility function. The second direction supports the practicality of learning from a utility function that passes our first three sanity checks. Specifically, in the work we reviewed, the challenges of exploration and reward sparsity were partially addressed by reward shaping and low penalties for collisions. One could empirically demonstrate that, while learning from a reward function that passes the corresponding sanity checks, these challenges can instead be addressed by other methods. Third, the broad application of these sanity checks across numerous tasks would likely lead to further insights and refinement of the sanity checks. Fourth, one could develop more comprehensive methods for evaluating utility functions with respect to human stakeholders' interests. Fifth, trial-and-error reward design is common in AD and, if we extrapolate from informal conversations with other researchers, is currently the dominant form of reward design across RL tasks in general. Investigation of the consequences of this ad hoc technique would shed light on an unsanctioned yet highly impactful practice. Finally, the community would benefit from research that constructs best practices for the manual design of reward functions for arbitrary tasks.

\section*{Acknowledgements}
We thank Ishan Durugkar, Garrett Warnell, Sam Devlin, Yunshu Du, James MacGlashan, and Brian Christian for their valuable feedback. We also thank the authors of the focus papers who generously answered our questions about their publications' details. This work is a collaboration between Bosch and the the Learning Agents Research Group (LARG) at UT Austin. LARG research is supported in part by NSF (CPS-1739964, IIS-1724157, FAIN-2019844), ONR (N00014-18-2243), ARO (W911NF-19-2-0333), DARPA, Lockheed Martin, GM, Bosch, and UT Austin's Good Systems grand challenge. Peter Stone serves as the Executive Director of Sony AI America and receives financial compensation for this work. The terms of this arrangement have been reviewed and approved by the University of Texas at Austin in accordance with its policy on objectivity in research.

 \bibliographystyle{elsarticle-harv}
\bibliography{references}

\newpage

\appendix

\section{Reward functions in the 10 focus papers}
\label{app:rewfcns}

In this Appendix section, we describe reward functions and other problem specification details from the 10 papers that we evaluate in Section~\ref{sec:tools}. We also closely examined 9 other publications, but because the patterns in the 10 focus papers below were so consistent, we did not pursue clarifying these 9 additional publications' problem specifications sufficiently for us to be able to characterize them with confidence at the same level of detail as the focus papers described below.

In this section, papers are listed alphabetically by first author's last name. A $\dagger$ marks information obtained in part or fully through correspondence with the paper's author. Additionally, we include time limit information but do not typically report whether the RL agent is updated with a terminal transition when time expires and an episode is stopped, since we rarely have such information. However, we suspect that most of these time limits are meant to make training feasible and are not actually part of the problem specification, and we further suspect that agents are correctly not updated with a terminal transition upon time exhaustion.

\subsection{LeTS-Drive: Driving in a Crowd by Learning from Tree Search~\cite{cai2019lets}}

\paragraph{Reward function}
The reward function is the unweighted sum of 3 attributes, with the authors' stated purpose of each in brackets:
\begin{itemize}
    \item \relax [efficiency] $-0.1$ at each non-terminal time step (and $0$ for a terminal transition);
    \item \relax [smoothness] $-0.1$ if the action includes non-zero acceleration ($0$ otherwise); and
    \item \relax [safety] $-1000 \times (v^2+0.5)$ upon collision with a pedestrian or a static obstacle, where $v$ is the driving speed in m/s\authemail ~($0$ otherwise).
\end{itemize}

\paragraph{Time step duration}
Time steps are $100$ ms.

\paragraph{Discount factor} 
The discount factor $\gamma=1$.\authemail

\paragraph{Episodic/continuing, time limit, and termination criteria} 
The task is episodic. Regarding the time limit, episodes are computationally stopped after $120$ seconds (in simulation time) / $1200$ time steps, which is used for calculating the success rate. However, if a trajectory is stopped at $120$ seconds, none of the trajectory is used to update the value function, making it somewhat optimistic.\authemail~Termination criteria are collisions and the agent reaching the goal.\authemail

\subsection{Model-free Deep Reinforcement Learning for Urban Autonomous Driving~\cite{chen2019model} }

\paragraph{Reward function}
The reward function is the unweighted sum of 5 attributes:
\begin{itemize}
    \item $min(\text{speed}, 10 - \text{speed})$, the ego vehicle's speed in m/s, with a penalty for going too fast; 
    \item $-0.5 * (\text{steering angle})^2$, a penalty for the magnitude of the steering angle in radians; 
    \item $-10$ upon a collision ($0$ otherwise);
    \item $-1$ upon leaving the lane, incurred if the distance between the ego vehicle's center and the closest point on the provided route's polyline is greater than 2m ($0$ otherwise);\authemail and
    \item $-0.1$ every step to encourage quickly reaching the goal.
\end{itemize}

Reward is calculated for every 100 ms CARLA time step, but it is received by the agent only at its decision points, every 400 ms. That received reward is the sum of the four 100 ms rewards.

\paragraph{Time step duration}
Time steps are effectively 400 ms because ``frame skip'' is used to keep the chosen action unchanged for 4 frames, each of which is 100 ms (as in other research within the CARLA simulator).

\paragraph{Discount factor} 
The discount factor $\gamma=0.99$ and was applied at every decision point (i.e., every 400 ms).\authemail

\paragraph{Episodic/continuing, time limit, and termination criteria} 
The task is episodic. The time limit is $500$ time steps (i.e., $50$ s).\authemail ~
Termination criteria are getting to the goal state, collisions, leaving the lane, and running out of time.\authemail

\subsection{CARLA: An open urban driving simulator~\cite{dosovitskiy2017carla}}

\paragraph{Reward function}
The reward function is the weighted sum of 5 attributes:

$r = (1) \Delta d + (0.05) \Delta v + (-0.00002) \Delta c + (-2)\Delta s + (-2) \Delta o$.

These attributes are

\begin{itemize}
    \item $\Delta d$, the change in distance traveled in meters along the shortest path from start to goal, regularly calculated using the ego's current position;
    \item $\Delta v$, the change in speed in km/h;
    \item $\Delta c$, the change in collision damage (expressed in range $[0,1]$);
    \item $\Delta s$, the change in the proportion of the ego vehicle that currently overlaps with the sidewalk; and
    \item $\Delta o$, the change in the proportion of the ego vehicle that currently overlaps with the other lane.
\end{itemize}

\paragraph{Time step duration}
Time step information is not described in the paper, but $0.1$ s is the time step duration reported by all other papers that use the CARLA simulator and report their time step duration.

\paragraph{Discount factor} The first author suspects $\gamma=0.99$ (which is typical for A3C).\authemail

\paragraph{Episodic/continuing, time limit, and termination criteria} 
The task is episodic. The time limit for an episode is the amount of time it would take a car to follow the shortest path from the start state to a goal state, driving at 10 km/h. Termination occurs upon collision or reaching the goal. %

\subsection{Dynamic Input for Deep Reinforcement Learning in Autonomous Driving~\cite{huegle2019dynamic}}

\paragraph{Reward function}
The reward function is the unweighted sum of 3 attributes:
\begin{itemize}
    \item $1$ given every step;
    \item $-\frac{|v - v_{desired}|}{ v_{desired}}$; and
    \item $-0.01$ if the current action is a lane change (0 otherwise).
\end{itemize}

\paragraph{Time step duration}
Time steps for the agent are 2 seconds.%

\paragraph{Discount factor} 
The discount factor $\gamma=0.99$.\authemail

\paragraph{Episodic/continuing, time limit, and termination criteria} 
The task is continuing.\authemail\\
There is no time limit nor termination criterion.\authemail

\subsection{Navigating Occluded Intersections with Autonomous Vehicles using Deep Reinforcement Learning~\cite{isele2018navigating}}

\paragraph{Reward function}
The reward function is the unweighted sum of 3 attributes:
\begin{itemize}
    \item $-0.01$ given every step;
    \item $-10$ if a collision occurred (and $0$ otherwise); and
    \item $+1$ when the agent successfully reaches the destination beyond the intersection (and $0$ otherwise).
\end{itemize}

\paragraph{Time step duration}
Time steps are 200 ms, though some actions last 2, 4, or 8 time steps (i.e., the agent may frame skip).

\paragraph{Discount factor} 
The discount factor $\gamma=0.99$.\authemail

\paragraph{Episodic/continuing, time limit, and termination criteria} 
The task is episodic. In unoccluded scenarios, episodes are limited to 20 s. In occluded scenarios, episodes are limited to 60 s. Termination occurs upon success, running out of time, or collision. %

\subsection{End-to-End Race Driving with Deep Reinforcement Learning~\cite{jaritz2018end}}

\paragraph{Reward function}
Four different reward functions are evaluated, three of which are new and intentionally add reward shaping:
\begin{itemize}
    \item $r = v$
    \item $r = v(\cos{\alpha} - d)$ (``Ours'')
    \item $r = v(\cos{\alpha} - \max(d-0.5w,0))$ (``Ours - w/ margin'')
    \item $r = v(\cos{\alpha} - \frac{1}{1+\exp{-4(|d|-0.5w}})$ (``Ours - sigmoid'')
\end{itemize}

Above, $v$ is the vehicle velocity component in the direction of the lane's center line, $d$ is the distance from the middle of the road, $\alpha$ is the difference between the vehicle's heading and the lane's heading, and $w$ is the road width.

\paragraph{Time step duration}
Time step duration is 33ms, with 1 step per 30 FPS frame.\authemail 
~The game will pause to await the RL agent's action. 

\paragraph{Discount factor} 
The discount factor $\gamma=0.99$.\authemail~%

\paragraph{Episodic/continuing, time limit, and termination criteria} 
The task is episodic. No time limit is enforced. The termination criteria are the vehicle stops progressing or the vehicle goes off-road or the wrong direction.%

\subsection{CIRL: Controllable Imitative Reinforcement Learning for Vision-based Self-driving~\cite{liang2018cirl}}

\paragraph{Reward function}
The reward function is the unweighted sum of 5 attributes:%
\begin{itemize}[noitemsep,nolistsep]
    \item penalty for steering angles in ranges assumed incorrect for current command (e.g., going left during a turn-right command);
    \item speed (km/h), with penalties for going too fast on a turn and limits to speed-based reward when not turning (to keep under a speed limit);
    \item -100 upon collision with vehicles or pedestrians and -50 upon collision with anything else (e.g., trees and poles) (and $0$ otherwise);
    \item -100 for overlapping with the sidewalk (and $0$ otherwise); and
    \item -100 for overlapping with the opposite-direction lane (and $0$ otherwise).
\end{itemize}

\paragraph{Time step duration}
Time steps last 100 ms, the same as has been reported in other research conducted within the CARLA simulator.

\paragraph{Discount factor} 
The discount factor $\gamma=0.9$.

\paragraph{Episodic/continuing, time limit, and termination criteria} 
The task is episodic. The time allotted for an episode is the amount of time to follow the ``optimal path'' to the goal at 10 km/h. Termination occurs upon successfully reaching the destination, having a collision, or exhausting the allotted time.

\subsection{Deep Distributional Reinforcement Learning Based High-Level Driving Policy Determination~\cite{min2019deep}}

\paragraph{Reward function}
The reward function is the unweighted sum of 4 attributes:
\begin{itemize}
    \item $\frac{v-40}{40}$, where $v$ is speed in km per hour within the allowed range of $[40, 80]$ km/h;
    \item $0.5$ if the ego vehicle overtakes another vehicle ($0$ otherwise);
    \item $-0.25$ if the ego vehicle changes lane ($0$ otherwise); and
    \item $-10$ if the ego vehicle collides ($0$ otherwise).
\end{itemize}

\paragraph{Time step duration}
The time step duration is the time between frames in the Unity-based simulator. The correspondence of such a frame to seconds in the simulated world was unknown to the first author.\authemail

\paragraph{Discount factor} 
The discount factor $\gamma=0.99$.

\paragraph{Episodic/continuing, time limit, and termination criteria} 
The task is episodic. There is no time limit.\authemail~ Termination occurs upon collision with another vehicle or when the ego vehicle travels the full track length (2500 Unity spatial units\authemail ), effectively reaching a goal. %

\subsection{Learning hierarchical behavior and motion planning for autonomous driving~\cite{wang2020learning}}

\paragraph{Reward function}
The reward function is defined separately for transitions to terminal and non-terminal states. For transitions to terminal states, reward is one of the following:
\begin{itemize}
    \item $100$ if the goal was reached;
    \item $-50$ upon a collision or running out of time;
    \item $-10$ for a red light violation; or
    \item $-1$ if the ego vehicle is in the wrong lane.
\end{itemize}

The reward for a single \emph{non-terminal} high-level behavioral step is the negative sum of the costs of the shorter steps within the best trajectory found by the motion planner, which executes the high-level action. Expressed as the additive inverse of cost, this reward for a high-level behavioral step is the unweighted sum of these 3 attributes:
\begin{itemize}
    \item $\frac{- \sum_t{t^2|v_{ref} - v(t)|}}{\sum_t{t^2}}$, which rewards speeds close to the desired speed, $v_{ref}$;
    \item $\frac{-1}{1 + \sum_t{|v(t)|}}$, which rewards based on distance traveled; and
    \item $\sum_t{[0.02 \times d_{olon}(t) + 0.1 \times d_{olat}(t)]}$, which rewards keeping larger distances from obstacles.
\end{itemize}

Distances are in meters and speed in m/s.\authemail ~Desired speed $v_{ref}$ is determined by the RL agent's high-level actions \emph{speed\_up} and \emph{speed\_down}, which each change $v_{ref}$ by a fixed percentage.
$v_{t}$ is the current speed. Time $t$ is a count from 1 over the time steps of the planned trajectory.\authemail~ $d_{olon}$ and $d_{olat}$ are respectively the longitudinal and lateral distance from the closest vehicle within 20 meters, measured between the centers of the cars. If no vehicle is present within that range, $d_{olon}=20$ and $d_{olat}=3.5$.\authemail

\paragraph{Time step duration}
A new high-level action is only chosen by the RL agent once the planned trajectory from the previous high-level action is completed. In the simulator, this results in an \emph{average} time step of 1s for the RL-based behavior policy. The the lower-level motion planner averages 100ms per planning iteration.

\paragraph{Discount factor} 
Discount factor $\gamma=0.99$.

\paragraph{Episodic/continuing, time limit, and termination criteria} 
The task is episodic. The time limit is the time required to travel the length of a randomly generated route in a CARLA town at 10 km/h. Episodes are terminated when the goal is reached or any of the following occur: a collision, driving out of the lane, or a red light violation.

\subsection{End-to-End Model-Free Reinforcement Learning for Urban Driving using Implicit Affordances~\cite{toromanoff2020end}}

\paragraph{Reward function}

The reward function outputs $r = r_{speed} + 0.5 \times r_{dist} + 0.5 \times r_{heading}$ and is in $[-1,1]$.
\begin{itemize}
    \item $r_{speed}$: $1 - |s_{desired} - s_{ego}| / 40$,\authemail~ an attribute in $[0,1]$ ($40$ km/h is the maximum $s_{desired}$\authemail) that is inversely proportional to the absolute difference in the actual speed and the desired speed; 
    \item $r_{dist}$: $- d_{path} / d_{max}$, an attribute in $[-1,0]$, where $d_{max}=2.0$ and $d_{path}$ is the distance in meters from the closest point on the optimal path's spline;
    \item $r_{heading}$: ${clip}(-1,0,b \times |\theta_{ego} - \theta_{path}|)$\authemail, an attribute in $[-1,0]$, where $\theta_{ego}$ is the ego vehicle's heading and $\theta_{path}$ is the heading of the optimal path spline at its closest point%
    ; and
    \item $-1$ upon termination ($0$ otherwise).
\end{itemize}

To determine reward, an \emph{optimal path} is created via the waypoint API. This path is optimal with simplifying assumptions. In training, this waypoint-generated optimal path is generated by randomly choosing turn directions at intersections.

The unit for speeds is km/h, and the unit for angles is degrees. The desired speed, $s_{desired}$, is hard coded based on the presence of traffic lights and obstacles, using ``priveleged'' information available only during training. $d_{max}$ is half the width of a lane (which apparently is always $2$ m in CARLA). In the formula for $r_{heading}$, $b = (-1/10)$ when the optimal path's accompanying high-level command is to follow the lane or go straight through an intersection, and $b = (-1/25)$ when the command is to turn left or right through an intersection.\authemail

We note that this reward function appears somewhat supervisory, teaching the target speed in various contexts and to stay close to the precomputed path. Further, the context required to calculate reward is not available to the agent during testing.

\paragraph{Time step duration}
Time steps last $100$ ms, the same as has been reported in other research conducted within the CARLA simulator.

\paragraph{Discount factor} 
Discount factor $\gamma=0.99$.\authemail

\paragraph{Episodic/continuing, time limit, and termination criteria} 
The task is episodic. There is no time limit during training. During training, there were no successful terminations (i.e., upon reaching a destination); instead, driving continued along a procedurally generated route until an undesirable termination condition was met.\authemail~
Termination conditions are when the agent is further from the optimal path than $d_{max}$, collisions, running a red light, and having $0$ speed when neither behind an obstacle nor waiting at a red traffic light.

\section{Reward shaping examples per paper}
\label{app:shaping}

Of the 19 papers we reviewed, we are highly confident that 13 include reward shaping. Below is at least one example per paper of behavior discouraged or encouraged by reward shaping. The following examples are discouraged behavior, penalized via negative reward:
\begin{itemize}
    \item deviating from the center of the lane~\citep{jaritz2018end,toromanoff2020end,paxton2017combining,tang2019towards}, 
    \item changing lanes~\citep{huegle2019dynamic},  
    \item overlapping with the opposite-direction lane~\citep{dosovitskiy2017carla,liang2018cirl}, 
    \item overlapping a lane boundary~\citep{henaff2019model},
    \item delaying entering the intersection when it is the ego vehicle's right of way at a stop sign~\citep{li2019urban},
    \item deviating from steering straight~\citep{chen2019model},
    \item side-ways drifting (on a race track)~\citep{liu2017learning},
    \item getting close to other vehicles~\citep{henaff2019model}, and
    \item having the turn signal on~\citep{tang2019towards}.
\end{itemize} 
These following two examples are behaviors that are encouraged by positive reward:
\begin{itemize}
    \item passing other vehicles~\citep{min2019deep} and  
    \item increasing distances from other vehicles~\citep{wang2020learning}.
\end{itemize}

Additionally, 2 papers had reward attributes that were somewhat defensible as not constituting reward shaping. The reward function in \citet{wang2019quadratic} also includes a penalty correlated with the lateral distance from center of the lane, but their reinforcement learning algorithm is \emph{explicitly} a lower-level module that receives a command for which lane to be in, and one could argue the subtask of this module is to stay inside that lane. However, being perfectly centered in the lane is not part of that high-level command, so we think the argument for considering it to be reward shaping is stronger than the argument for the alternative. 
A reward attribute in \citet{aradi2018policy} encourages being in the rightward lane if no car is in the way, which fits laws in~\emph{some} US states that require drivers to keep right unless passing. Therefore, whether their reward function includes reward shaping depends on the laws for the location of the ego vehicle.

Lastly, we are confident that 4 papers do not include reward shaping~\citep{isele2018navigating,cai2019lets,kendall2019learning,mirchevska2018high}.

\section{Calculation of trajectory returns}
\label{app:trajcalcs}

This appendix section describes how we estimate the return for various trajectories under each reward function. These calculations are used for two related sanity checks for reward functions: comparing preference orderings (Section~\ref{sec:preforder}) and comparing indifference points (Section~\ref{sec:indifference}).

Recall that we estimate returns for 3 different types of trajectories:
\begin{itemize}
    \item $\tau_{crash}$, a drive that is successful until crashing halfway to its destination;
    \item $\tau_{idle}$, the safe trajectory of a vehicle choosing to stay motionless where it was last parked; and
    \item $\tau_{succ}$, a trajectory that successfully reaches the destination.
\end{itemize}

We additionally remind readers that an indifference point is calculated by solving the following equation for $p$ when $\tau_A \prec \tau_B \prec \tau_C$: 
\begin{equation*}
G(\tau_B) =
p G(\tau_C) + (1-p)G(\tau_A).
\end{equation*}
In this paper, we calculate it specifically with $G(\tau_{idle}) =
p G(\tau_{succ}) + (1-p)G(\tau_{crash})$. And from $p$, the \emph{safety indifference point}, expressed in km per crash, can be calculated as $((p/(1-p))+0.5) \times \text{path length}$, where \emph{path length} is the length of a successful trajectory. Note that $p/(1-p)$ expresses the amount of $\tau_{succ}$ trajectories per half-length $\tau_{crash}$ trajectory at the indifference point, making $(p/(1-p))+0.5$ the path lengths driven per collision. 

To illustrate, assume that for successful-until-collision $\tau_{crash}$, $G(\tau_{crash}) = -10$; for motionless $\tau_{idle}$, $G(\tau_{idle}) = -5$; and for successful $\tau_{succ}$, $G(\tau_{succ}) = 10$. Recall that the indifference point $p$ is where the utility function has no preference over $\tau_{idle}$ and a lottery between $\tau_{crash}$ and $\tau_{succ}$ according to $p G(\tau_{succ}) + (1-p) G(\tau_{crash})$. For this example, $-5 = (p \times 10) + ((1-p) \times -10)$, and solving the equation results in $p = 0.25$. Therefore, the utility function would prefer driving (more than not driving) with any ratio higher than 1 success to 3 crashes. Let us also assume that a path length is $1$ km. Therefore, at the indifference point, for each collision the car would drive $0.33$ km on successful drives and \emph{half} of 1 km on a drive with a collision, which is
\begin{align*}
(\frac{p}{1-p}+0.5 \text{ \textscript{path lengths per collision}}) \times  1 \text{\textscript{ km per path length}} &= \frac{0.25}{0.75}+0.5 \\
&= 0.83 \text{\textscript{ km per collision}}.
\end{align*} %
Calculations of safety indifference points are given below for the two papers for which $\tau_{crash} \prec \tau_{idle} \prec \tau_{succ}$~\cite{cai2019lets,isele2018navigating}.

In our descriptions below, we try to name every significant assumption we make. Readers might find it useful to choose different assumptions than ours to test how sensitive our analysis is to these assumptions; we expect the reader will find no changes in the qualitative results that arise from this quantitative analysis.

\subsection{General methodology and assumptions} 
\textbf{To estimate return for some trajectory $\tau_i$, we do not fully ground it as a sequence of state-action pairs.} As we show in the remainder of this appendix section, the exact state-action sequence is not needed for estimating certain trajectories' returns under these reward functions. 

For calculating reward per time step, we adhere to the following methodology. 
\begin{itemize}
    \item To determine the return/utility for a successful portion of a trajectory, we assume no unnecessary penalties are incurred (e.g., driving on the sidewalk). 
    \item For positive attributes of reward, we choose the value that gives the maximum outcome. If the maximum is unclear, we choose outcomes for attributes that are as good or better than their best-reported experimental results. Lastly, if experimental results do not include a measure of the reward attribute, we attempt to calculate a value that is better than what we expect a typical human driver to do. 
    \item Path lengths are the given length of the paper's driving tasks or our estimation of it. If the given information is insufficient to estimate the path length, we assume it to be 1 km.
\end{itemize}

For the papers paper below, we write out the units for each term in the equations for the return of a trajectory. We encourage the reader to refer to the paper's corresponding subsection in \ref{app:rewfcns} to understand our calculations of return. Additionally, to aid the reader, we use specific colors for terms expressing the \textcolor{purple}{time limit}, \textcolor{Goldenrod}{path length}, \textcolor{orange}{time step duration}, and \textcolor{OliveGreen}{speed}. For reward functions that are sums of attributes, in our return calculations we maintain the order of the attributes as they were described in \ref{app:rewfcns}, and we include $0$ terms for attributes that do not affect return for the corresponding trajectory.  %

\subsection{Assumptions and calculations for each paper}

\paragraph{LeTS-Drive: Driving in a Crowd by Learning from Tree Search~\cite{cai2019lets}}

Driving maps are 40 m $\times$ 40 m and are each a single intersection or curve, and from this map size we assume the path length is 40 m. This assumption appears reasonable because the car is spawned in a random location.

For the successful path $\tau_{succ}$ and the successful portion of path $\tau_{crash}$, we use mean task time and the number of deceleration events reported for their best algorithm. The motionless $\tau_{idle}$ involves time running out, which is not actually a termination event in their method but is treated so here. The driving speed when collision occurs is assumed to be the mean speed for LeTS-Drive calculated from their Table 1's Time-to-goal and the path length.

Path length: $0.04$ km.

A trajectory that is successful until collision:
\begin{equation*}
\begin{split}
G(\tau_{crash}) &= -515.71 \\
&= (-0.1 \textscript{reward / time step} \times \textcolor{orange}{10 \textscript{time step / s}} \times 29.6 \textscript{s / successful trajectory} \\
& \quad \quad \times 0.5 \textscript{ of full path length traveled}) \\ 
& \quad + (-0.1 \textscript{reward / acceleration event} \times 18.2 \textscript{mean acceleration events / successful episode}\\
& \quad \quad \times 0.5 \textscript{ of full path length traveled}) \\
& \quad + (-1000 \times (\textcolor{Goldenrod}{40 \textscript{m path length}} / 29.6 \textscript{s / successful trajectory})^2+0.5)
\end{split}
\end{equation*}

A motionless trajectory:
\begin{equation*}
\begin{split}
G(\tau_{idle}) &= -120 \\
&= (-0.1 \textscript{efficiency reward / time step} \times 
\textcolor{orange}{10 \textscript{time step / s}} \times 
\textcolor{purple}{120 \textscript{s before time limit is reached}}) \\ 
& \quad + (-0.1 \textscript{smoothness reward / acceleration event} \times 0 \textscript{acceleration events / episode}) \\
& \quad + 0 \textscript{safety reward}
\end{split}
\end{equation*}

A successful trajectory:
\begin{equation*}
\begin{split}
G(\tau_{succ}) &= -31.42 \\ 
&= (-0.1 \textscript{efficiency reward / time step} \times \textcolor{orange}{10 \textscript{time step / s}} \times 29.6 \textscript{s / successful trajectory}) \\ 
& \quad + (-0.1 \textscript{smoothness reward / acceleration event} \\
& \quad \quad \times 18.2 \textscript{mean acceleration events / successful episode}) \\
& \quad + (0 \textscript{for no collision})
\end{split}
\end{equation*}

Calculation of the indifference point:
\begin{equation*}
\begin{split}
G(\tau_{idle}) &= p G(\tau_{succ}) + (1-p) G(\tau_{crash}) \\
p &= \frac{G(\tau_{idle}) - G(\tau_{crash})}{G(\tau_{succ}) - G(\tau_{crash})} \\
p &= \frac{-120 - (-515.71)}{-31.42 - (-515.71)} \\
p &= 0.9617
\end{split}
\end{equation*}

Calculation of km per collision at the indifference point:
\begin{equation*}
\begin{split}
(\frac{p}{1-p}+0.5 \textscript{ path lengths per collision}) \times  0.04 \textscript{ km per path length} 
&= (\frac{0.9617}{1 - 0.9617} + 0.5) \\
& \quad \times 0.04 \\
&= 1.02 \textscript{ km per collision}
\end{split}
\end{equation*}

\paragraph{Model-free Deep Reinforcement Learning for Urban Autonomous Driving~\cite{chen2019model} }

We assume a constant speed of $5$ m/s ($18$ km/h), which is the most reward-giving speed for the speed-based reward attribute. This paper focuses on a specific roundabout navigation task, which presumably would be a shorter route than those used in the more common CARLA benchmarks first established by~\citet{dosovitskiy2017carla}. Accordingly, differing from our assumptions for most other CARLA evaluations, we assume a $0.125$ km path length, the distance which can be achieved at the above speed in exactly half of the permitted $50$ s time limit. 
We further assume that the steering angle is always $0$ (avoiding a penalty) and that the agent never leaves its lane except upon a collision. Also, recall that reward is accrued at $100$ ms time steps (whereas discounting is applied at $400$ ms time steps). %

Path length: $0.125$ km.

A trajectory that is successful until collision:
\begin{equation*}
\begin{split}
G(\tau_{crash}) &= 601.5  \\ 
& =(\textcolor{OliveGreen}{5 \textscript{ m / s}} \\
& \quad \quad \times (\frac{\textcolor{Goldenrod}{125\textscript{ m path length}} \times 0.5 \textscript{ of full path length traveled}}{ \textcolor{OliveGreen}{5\textscript{ m / s}}} \times \textcolor{orange}{10 \textscript{~time step / s}})) \\
& \quad + (0 \textscript{for no deviations from 0 steering angle}) + (-10 \textscript{for the collision}) \\
& \quad + (-1 \textscript{ for leaving the lane before the collision}) \\
& \quad + (-0.1 \textscript{ constant reward / time step} \\
& \quad \quad \times (\frac{\textcolor{Goldenrod}{125\textscript{ m path length}} \times 0.5 \textscript{ of full path length traveled}}{ \textcolor{OliveGreen}{5\textscript{ m / s}}} \times \textcolor{orange}{10 \textscript{~time step / s}})) \\
\end{split}
\end{equation*}

A motionless trajectory:
\begin{equation*}
\begin{split}
G(\tau_{idle}) &= -50.0 \\
&= (0 \textscript{for 0 speed throughout}) \\
& \quad + (0 \textscript{for no deviations from 0 steering angle}) + (0 \textscript{for no collision}) \\
& \quad + (0 \textscript{ for never leaving the lane}) \\
& \quad + (-0.1 \textscript{ constant reward / time step} \times (\textcolor{purple}{50 \textscript{s time limit}} \times \textcolor{orange}{10 \textscript{~time step / s}})
\end{split}
\end{equation*}

A successful trajectory:
\begin{equation*}
\begin{split}
G(\tau_{succ}) &= 1225.0   \\ 
& =(\textcolor{OliveGreen}{5 \textscript{ m / s}} \times (\frac{\textcolor{Goldenrod}{125\textscript{ m path length}}}{ \textcolor{OliveGreen}{5\textscript{ m / s}}} \times \textcolor{orange}{10 \textscript{~time step / s}})) \\
& \quad + (0 \textscript{for no deviations from 0 steering angle}) + (0 \textscript{for no collision})~ \\
& \quad + (0 \textscript{ for never leaving the lane}) \\
& \quad + (-0.1 \textscript{ constant reward / time step} \\
& \quad \quad \times (\frac{\textcolor{Goldenrod}{125\textscript{ m path length}}} {\textcolor{OliveGreen}{5\textscript{ m / s}}} \times \textcolor{orange}{10 \textscript{~time step / s}})) \\
\end{split}
\end{equation*}

\paragraph{CARLA: An open urban driving simulator~\cite{dosovitskiy2017carla}}

For a successful drive, we assume a $1$ km path, a change in speed from start to finish is $60$ km/h, and that no overlap occurs with sidewalk or other lane. For a trajectory with a collision, we assume the collision damage is total (i.e., 1) and the ego vehicle completely overlaps with sidewalk or other lane at $60$ km/h. %

Path length: $1$ km.

A trajectory that is successful until collision:
\begin{equation*}
\begin{split}
G(\tau_{crash}) &= 501.00 \\
& = (\textcolor{Goldenrod}{500 \textscript{ m path length}} \times 1 \textscript{reward / m traveled}) \\
& \quad + (0.05 \times \textcolor{OliveGreen}{60 \textscript{km / h increased over trajectory}})  \\
& \quad + (-0.00002 \times 1 \textscript{ for full collision damage}) \\
& \quad + (0 \textscript{ for never overlapping with the sidewalk}) \\
& \quad + (-2 \times 1 \textscript{ for fully leaving the lane before collision}) \\ 
&
\end{split}
\end{equation*}

A motionless trajectory:
\begin{equation*}
\begin{split}
G(\tau_{idle}) &= 0 \\
& = (0 \textscript{ for no distance traveled})+ (\textcolor{OliveGreen}{0 \textscript{km / h increased over trajectory}}) \\
& \quad + (0 \textscript{ for no collision damage}) + (0 \textscript{ for never overlapping with the sidewalk})  \\
& \quad + (0 \textscript{ for never leaving the lane})
\end{split}
\end{equation*}

A successful trajectory:
\begin{equation*}
\begin{split}
G(\tau_{succ}) &= 1003 \\ 
& = (\textcolor{Goldenrod}{1000 \textscript{ m path length}} \times 1 \textscript{reward / m traveled}) \\
& \quad + (0.05 \times \textcolor{OliveGreen}{60 \textscript{km / h increased over trajectory}}) \\
& \quad + (0 \textscript{ for no collision damage}) + (0 \textscript{ for never overlapping with the sidewalk})  \\
& \quad + (0 \textscript{ for never leaving the lane})
\end{split}
\end{equation*}

\paragraph{Dynamic Input for Deep Reinforcement Learning in Autonomous Driving~\cite{huegle2019dynamic}}

Because a collision appears impossible in this task, this reward function was not involved in the analysis of preference orderings and indifference points.

\paragraph{Navigating Occluded Intersections with Autonomous Vehicles using Deep Reinforcement Learning~\cite{isele2018navigating}}

We assume that the path is $6$ lane widths, which is roughly what the "Left2" turn requires and that each lane is $3.35$ m wide (based on 11 feet appearing common enough, e.g., in \url{https://mutcd.fhwa.dot.gov/rpt/tcstoll/chapter443.htm}). For the successful drive, we assume $4$ s was required. We focus on the unoccluded scenario with a 20 s time limit.

Path length: $ (3.35 \textscript{ m / lane width} \times 6 \textscript{ lane widths / path length}) / 1000 \textscript{ m / km} \approx 0.02 $ km. 

A trajectory that is successful until collision:
\begin{equation*}
\begin{split}
G(\tau_{crash}) &= -10.1 \\
& = (0.5 \textscript{ of full path length traveled} \times (4 \textscript{s / successful trajectory} / \textcolor{orange}{0.2 \textscript{s / time step}}) \\
& \quad \quad \times -0.01 \textscript{ reward / time step}) \\
& \quad + (-10 \textscript{ for collision}) \\
& \quad + (0 \textscript{ for not reaching the goal})
\end{split}
\end{equation*}

A motionless trajectory:
\begin{equation*}
\begin{split}
G(\tau_{idle}) &= -1 \\
& = ((\textcolor{purple}{20 \textscript{s time limit}} / \textcolor{orange}{0.2 \textscript{s / time step}}) \times -0.01 \textscript{ reward / time step}) \\
& \quad + (0 \textscript{ for no collision}) \\
& \quad + (0 \textscript{ for not reaching the goal})
\end{split}
\end{equation*}

A successful trajectory:
\begin{equation*}
\begin{split}
G(\tau_{succ}) &= 0.8 \\ 
& = ((4 \textscript{s / successful trajectory} / \textcolor{orange}{0.2 \textscript{s / time step}}) \times -0.01 \textscript{ reward / time step}) \\
& \quad + (0 \textscript{ for no collision}) \\
& \quad + (1 \textscript{ for reaching the goal})
\end{split}
\end{equation*}

Calculation of the indifference point:
\begin{equation*}
\begin{split}
G(\tau_{idle}) &= p G(\tau_{succ}) + (1-p) G(\tau_{crash}) \\
p &= \frac{G(\tau_{idle}) - G(\tau_{crash})}{G(\tau_{succ}) - G(\tau_{crash})} \\
p &= \frac{-1 - (-10.1)}{0.8 - (-10.1)} \\
p &= 0.8349
\end{split}
\end{equation*}

Calculation of km per collision at the indifference point:
\begin{equation*}
\begin{split}
(\frac{p}{1-p}+0.5 \textscript{ path lengths per collision}) \times  \frac{0.02 \textscript{ km}}{\textscript{1 path length}}
&= (\frac{0.8349}{1 - 0.8349} + 0.5) \times 0.02 \\
&= 0.11 \textscript{ km per collision}
\end{split}
\end{equation*}

\paragraph{End-to-End Race Driving with Deep Reinforcement Learning~\cite{jaritz2018end}}

Note that the domain is a car-racing video game, so safety constraints differ from autonomous driving driving. For successful driving, we assume $72.88$ km/h, which is the average speed reported, and a $9.87$ km track. %
We assume that the ego vehicle's heading is always aligned with the lane and the car is always in the center of the lane.

Path length: $9.87$ km.

A trajectory that is successful until collision:
\begin{equation*}
\begin{split}
G(\tau_{crash}) &= 532980 \\
& = ((\textcolor{Goldenrod}{9.87 \textscript{km path length}} \times 
0.5 \textscript{ of full path length traveled} 
\times (1/\textcolor{OliveGreen}{72.88 \textscript{km / h}}) \\
& \quad \quad \quad \times 3600 \textscript{s / h}
\times \textcolor{orange}{30 \textscript{time steps / s}} 
)  \\
&  \quad \quad \times \textcolor{OliveGreen}{72.88 \textscript{km / h}} \times  \textcolor{OliveGreen}{1 \textscript{reward per km / h}}) \\
& \quad + (0 \textscript{for heading always aligned with the lane}) + (0 \textscript{for always at the lane center})
\end{split}
\end{equation*}

A motionless trajectory:
\begin{equation*}
\begin{split}
G(\tau_{idle}) &= 0 \\
& = (0 \textscript{for 0 km / h always}) \\
& \quad + (0 \textscript{for heading always aligned with the lane}) + (0 \textscript{for always at the lane center})
\end{split}
\end{equation*}

A successful trajectory:
\begin{equation*}
\begin{split}
G(\tau_{succ}) &= 1065960 \\ 
& = ((\textcolor{Goldenrod}{9.87 \textscript{km path length}}
\times (1/\textcolor{OliveGreen}{72.88 \textscript{km / h}})
\times 3600 \textscript{s / h}
\times \textcolor{orange}{30 \textscript{time steps / s}} 
)  \\
& \quad \quad \times \textcolor{OliveGreen}{72.88 \textscript{km / h}} \times  \textcolor{OliveGreen}{1 \textscript{reward per km / h}}) \\
& \quad + (0 \textscript{for heading always aligned with the lane}) + (0 \textscript{for always at the lane center})
\end{split}
\end{equation*}

\paragraph{CIRL: Controllable Imitative Reinforcement Learning for Vision-based Self-driving~\cite{liang2018cirl}}

For successful drives, we assume a $60$ km / h speed that is within speed limit. We assume no penalties are incurred. When a collision occurs, we assume overlap the opposite-direction lane for $1$ s (which has an equivalent impact as overlap with the sidewalk for $1$ s), a $60$ km / h speed that is within the speed limit, no steering-angle penalty, and collision with a vehicle specifically. As for other CARLA-based research, we assume a $1$ km successful trajectory, which creates a $6$ minute time limit.

Path length: $1$ km.

A trajectory that is successful until collision:
\begin{equation*}
\begin{split}
G(\tau_{crash}) &= 16900 \\
& = (0 \textscript{for always-$0$ steering angle}) \\
& \quad + (\textcolor{OliveGreen}{60 \textscript{km / h}} \\
& \quad \quad \times ((\textcolor{Goldenrod}{1 \textscript{km path length}} \times 0.5 \textscript{ of full path length traveled})
/ \textcolor{OliveGreen}{60 \textscript{km / h}}) 
\\
& \quad \quad \quad \times 3600 \textscript{s / h} \times \textcolor{orange}{10 \textscript{times steps / s}}) \\
& \quad + (-100 \textscript{for collision with a vehicle}) + (0 \textscript{ for never overlapping with the sidewalk}) \\
& \quad + (-100 \textscript{ for overlapping with the opposite-direction lane} \times 1 \textscript{s of overlap} \\
& \quad \quad \times \textcolor{orange}{10 \textscript{times steps / s}})
\end{split}
\end{equation*}

A motionless trajectory:
\begin{equation*}
\begin{split}
G(\tau_{idle}) &= 0 \\
& = (0 \textscript{for always-$0$ steering angle}) + (0 \textscript{for always \textcolor{OliveGreen}{$0$ km / h}}) \\
& \quad + (0 \textscript{for no collision}) + (0 \textscript{ for never overlapping with the sidewalk}) \\
& \quad + (0 \textscript{ for never overlapping with the opposite-direction lane})
\end{split}
\end{equation*}

A successful trajectory:
\begin{equation*}
\begin{split}
G(\tau_{succ}) &=  36000 \\ 
& = (0 \textscript{for always-$0$ steering angle}) \\
& \quad + (\textcolor{OliveGreen}{60 \textscript{km / h}} \\
& \quad \quad \times (\textcolor{Goldenrod}{1 \textscript{km path length}}
/ \textcolor{OliveGreen}{60 \textscript{km / h}}) 
\times 3600 \textscript{s / h} \times \textcolor{orange}{10 \textscript{times steps / s}}) \\
& \quad + (0 \textscript{for no collision}) + (0 \textscript{ for never overlapping with the sidewalk}) \\
& \quad + (0 \textscript{ for never overlapping with the opposite-direction lane})
\end{split}
\end{equation*}

\paragraph{Deep Distributional Reinforcement Learning Based High-Level Driving Policy Determination~\cite{min2019deep}}

For a successful trajectory and the successful portion of the trajectory with a collision, we assume $17$ overtakes per km (based on ``Distance'' in meters and ''Num overtake'' statistics shown in Fig. 7 in their paper and assuming those statistics are taken from a good trajectory) and an average of $1$ lane change per overtake. We also assume that the car is always driving at $80$ km / h, the speed that accrues the most reward. Since the minimum speed is $40$ km / h and stopping in this paper's task---highway driving---would be unsafe, we instead assume the vehicle can decline to be deployed for 0 return. Since the first author did not know the duration of a time step in simulator time, we assume a common Unity default of 30 frames per second (and therefore 30 time steps per second) and that Unity processing time equals simulator time; consequently, $0.033$ s time steps are assumed. We also assume a $1$ km path, since the first author also did not have access to the path length for the task.

Path length: $1$ km.

A trajectory that is successful until collision:
\begin{equation*}
\begin{split}
G(\tau_{crash}) &= 673.9 \\
& = (((\textcolor{OliveGreen}{80 \textscript{km / h}}-40)/40) \\
& \quad \quad \times (0.5 \textscript{ of full path length traveled} \times \frac{\textcolor{Goldenrod}{1 \textscript{km path length}}}{\textcolor{OliveGreen}{80 \textscript{km / h}}} \times 3600 \textscript{s / h} \\
& \quad \quad \quad \times \textcolor{orange}{30 \textscript{time steps / s}})) \\
& \quad + (0.5 \textscript{reward / overtake} \\
& \quad \quad \times (17 \textscript{overtakes / km} \times \textcolor{Goldenrod}{1 \textscript{km path length}} \times 0.5 \textscript{ of full path length traveled})) \\
& \quad + (-0.25 \textscript{reward / lane change} \\
& \quad \quad \times (17 \textscript{overtakes / km} \times 1 \textscript{lane change / overtake} \times \textcolor{Goldenrod}{1 \textscript{km path length}} \\
& \quad \quad \quad \times 0.5 \textscript{ of full path length traveled})) \\
& \quad + (-10 \textscript{for collision})
\end{split}
\end{equation*}

A motionless trajectory:
\begin{equation*}
\begin{split}
G(\tau_{idle}) &= 0 \\
\end{split}
\end{equation*}

A successful trajectory:
\begin{equation*}
\begin{split}
G(\tau_{succ}) &= 1357.9 \\ 
& = (\frac{\textcolor{OliveGreen}{80 \textscript{km / h}}-40}{40} \times (\frac{\textcolor{Goldenrod}{1 \textscript{km path length}}}{\textcolor{OliveGreen}{80 \textscript{km / h}}} \times 3600 \textscript{s / h} \times \textcolor{orange}{30 \textscript{time steps / s}})) \\
& \quad + (0.5 \textscript{reward / overtake} \times (17 \textscript{overtakes / km} \times \textcolor{Goldenrod}{1 \textscript{km path length}} )) \\
& \quad + (-0.25 \textscript{reward / lane change} \\
& \quad \quad \times (17 \textscript{overtakes / km} \times 1 \textscript{lane change / overtake} \times \textcolor{Goldenrod}{1 \textscript{km path length}} \times )) \\
& \quad + (0 \textscript{for no collision})
\end{split}
\end{equation*}

\paragraph{Learning hierarchical behavior and motion planning for autonomous driving~\cite{wang2020learning}}

We assume a $1$ km path length and a speed of $60$ km / h (or $16.67$ m / s), as we do for most other CARLA evaluations. We also assume that the desired speed $v_{ref}$ is always $60$ km / h, making the first per-time-step component result in 0 reward each step, and that high-level RL time steps last exactly $1$ s, which is their reported mean duration. Lastly, we assume that no other vehicles ever are closer than $20$ m from the ego vehicle. Because the path length is assumed to be $1$ km, the time limit is $360$ s (based on the time limit information in \ref{app:rewfcns}).

Path length: $1$ km.

A trajectory that is successful until collision:
\begin{equation*}
\begin{split}
G(\tau_{crash}) &= 174.8 \\
& = (0 \textscript{for always matching $v_{ref}$}) \\
& \quad +
((\frac{\textcolor{Goldenrod}{1 \textscript{km path length}} \times 0.5 \textscript{ of full path length traveled}}{ 
\textcolor{OliveGreen}{60 \textscript{km / h}}} 
\times 3600 \textscript{s / h} \\
& \quad \quad \quad \times \textcolor{orange}{1 \textscript{RL time steps / s}}) \\
& \quad \quad \times \frac{-1}{1 + (\textcolor{orange}{10 \textscript{planning time steps / RL time step}}  \times \textcolor{OliveGreen}{16.67 \textscript{m / s}}) }) \\
& \quad +
((\frac{\textcolor{Goldenrod}{1 \textscript{km path length}} \times 0.5 \textscript{ of full path length traveled}}{ 
\textcolor{OliveGreen}{60 \textscript{km / h}}} 
\times 3600 \textscript{s / h} \\
& \quad \quad \quad \times \textcolor{orange}{1 \textscript{RL time steps / s}}) \\
& \quad \quad \times \textcolor{orange}{10 \textscript{planning time steps / RL time step}} \times ((0.02 \times 20 \textscript{m}) + (0.1 \times 3.5 \textscript{m}))) \\
& \quad + (-50 \textscript{for a collision})
\end{split}
\end{equation*}

A motionless trajectory:
\begin{equation*}
\begin{split}
G(\tau_{idle}) &= -3711.2 \\
& =((\textcolor{purple}{360 \textscript{s time limit}} \times \textcolor{orange}{1 \textscript{RL time steps / s}}) \\
& \quad \quad \times (- |16.67 \textscript{m / s} - \textcolor{OliveGreen}{0 \textscript{m / s}}| \\
& \quad \quad \quad \times 1 \textscript{remaining terms after constant speed differential is moved outside summation}) \\
~
& \quad + ((\textcolor{purple}{360 \textscript{s time limit}} \times \textcolor{orange}{1 \textscript{RL time steps / s}}) \\
& \quad \quad \times (-1 / (1 + (\textcolor{orange}{10 \textscript{planning time steps / RL time step}} \times \textcolor{OliveGreen}{0 \textscript{km / h}})))) \\
~
&  \quad + ((\textcolor{purple}{360 \textscript{s time limit}} \times \textcolor{orange}{1 \textscript{RL time steps / s}}) \\
& \quad \quad \times \textcolor{orange}{10 \textscript{planning time steps / RL time step}}  \\
& \quad \quad \times ((0.02 \times 20 \textscript{m}) + (0.1 \times 3.5 \textscript{m}))) \\
~
~
& \quad + (-50 \textscript{for running out of time})
\end{split}
\end{equation*}

A successful trajectory:
\begin{equation*}
\begin{split}
G(\tau_{succ}) &= 549.6 \\ 
& = (0 \textscript{for always matching $v_{ref}$}) \\
& \quad +
((((\textcolor{Goldenrod}{1 \textscript{km path length}}) / 
\textcolor{OliveGreen}{60 \textscript{km / h}}) 
\times 3600 \textscript{s / h} 
\times \textcolor{orange}{1 \textscript{RL time steps / s}}) \\
& \quad \quad \times (-1 / (1 + (\textcolor{orange}{10 \textscript{planning time steps / RL time step}}  \times \textcolor{OliveGreen}{16.67 \textscript{m / s}} )))) \\
& \quad +
((((\textcolor{Goldenrod}{1 \textscript{km path length}}) / 
\textcolor{OliveGreen}{60 \textscript{km / h}}) 
\times 3600 \textscript{s / h} 
\times \textcolor{orange}{1 \textscript{RL time steps / s}}) \\
& \quad \quad \times \textcolor{orange}{10 \textscript{planning time steps / RL time step}} \\
& \quad \quad \times ((0.02 \times 20 \textscript{m}) + (0.1 \times 3.5 \textscript{m}))) \\
& \quad + (100 \textscript{for reaching the goal})
\end{split}
\end{equation*}

\paragraph{End-to-End Model-Free Reinforcement Learning for Urban Driving using Implicit Affordances~\cite{toromanoff2020end}}

As we do for other CARLA evaluations, we assume a $1$ km path length. We assume a speed of $30$ km / h, based on the first author's report that $40$ km/h was their maximum speed. We also assume termination occurs upon reaching some destination, which is only true during testing, since training involves driving until termination by failure. No additional reward is given at such successful termination. For $\tau_{crash}$ and $\tau_{succ}$, we assume that the ego vehicle is always moving at the speed, location, and heading. We also assume that the $0$-speed termination condition is not applied immediately but rather at $10$ s and later; this assumption is made to avoid terminating at the starting time step, when the vehicle is spawned with a speed near $0$ km / h.

Path length: $1$ km.

A trajectory that is successful until collision:
\begin{equation*}
\begin{split}
G(\tau_{crash}) &= 599 \\
& = (1 \textscript{reward for $s_{ego}$ always matching $s_{desired}$} \\ 
& \quad \quad \times 
(\textcolor{Goldenrod}{1 \textscript{km path length}}
\times 0.5 \textscript{ of full path length traveled}) \times
(1 / 
\textcolor{OliveGreen}{30 \textscript{km / h}})  \\
& \quad \quad \times 3600 \textscript{s / h} 
\times \textcolor{orange}{10 \textscript{time steps / s}}) \\
& \quad + (0 \textscript{for always $d_{path} = 0$}) \\
& \quad + (0 \textscript{for $\theta_{ego}$ always matching $\theta_{path}$}) \\
& \quad + (-1 \textscript{for collision})\\
\end{split}
\end{equation*}

A motionless trajectory:
\begin{equation*}
\begin{split}
G(\tau_{idle}) &= 25 \\
& = ((1 - \frac{|\textcolor{OliveGreen}{0 \textscript{km / h}} - 30 \textscript{km / h} ~s_{desired}|} {40})  \\
& \quad \quad \times
(\textcolor{Goldenrod}{10 \textscript{s before 0-speed termination condition is applied}} %
\times \textcolor{orange}{10 \textscript{time steps / s}})) \\
& \quad + (0 \textscript{for always $d_{path} = 0$}) \\
& \quad + (0 \textscript{for $\theta_{ego}$ always matching $\theta_{path}$}) \\
\end{split}
\end{equation*}

A successful trajectory:
\begin{equation*}
\begin{split}
G(\tau_{succ}) &= 1200 \\
& = (1 \textscript{reward for $s_{ego}$ always matching $s_{desired}$} \times
(\textcolor{Goldenrod}{1 \textscript{km path length}} / 
\textcolor{OliveGreen}{30 \textscript{km / h}}) \\
& \quad \quad \times 3600 \textscript{s / h} 
\times \textcolor{orange}{10 \textscript{time steps / s}}) \\
& \quad + (0 \textscript{for always $d_{path} = 0$}) \\
& \quad + (0 \textscript{for $\theta_{ego}$ always matching $\theta_{path}$}) \\
\end{split}
\end{equation*}

\end{document}